\documentclass{article}
\usepackage[accepted]{icml2021}
\RequirePackage[loading]{tracefnt}

\usepackage{microtype}
\usepackage{graphicx}
\usepackage{subfigure}
\usepackage{booktabs} % for professional tables

\usepackage{amsmath,amssymb,mathtools}
\usepackage{bm}
\usepackage{amsthm}
\usepackage{graphicx,color,wrapfig,epsfig,epstopdf}
\usepackage{enumitem}
\usepackage{algorithm,algorithmic}
\def\x{\bm{x}}
\def\g{\bm{g}}
\def\v{\bm{v}}
\def\bdelta{\bm{\delta}}
\def\m{\bm{m}}
\def\z{\bm{z}}
\def\bzeta{\bm{\zeta}}
\def\ti{{(i)}}

\def\C{\mathcal{C}}
\def\Cref{\ref}
\def\bzeta{\bm{\zeta}}

\def\OA{{{{1-bit Adam}}}}

\newcommand\numberthis{\addtocounter{equation}{1}\tag{\theequation}}
\newtheorem{theorem}{Theorem}%[section]
\newtheorem{lemma}{Lemma}%[Lemma]
\newtheorem{corollary}{Corollary}
\newtheorem{assumption}{Assumption}

\begin{document}
\twocolumn[
\icmltitle{{\OA}: Communication Efficient Large-Scale Training with {Adam}'s Convergence Speed}
% \date{}
% \maketitle
\begin{icmlauthorlist}
\icmlauthor{Hanlin Tang}{mic,ur}
\icmlauthor{Shaoduo Gan}{eth}
\icmlauthor{Ammar Ahmad Awan}{mic}
\icmlauthor{Samyam Rajbhandari}{mic}
\icmlauthor{Conglong Li}{mic}
\icmlauthor{Xiangru Lian}{ur}
\icmlauthor{Ji Liu}{ur}
\icmlauthor{Ce Zhang}{eth}
\icmlauthor{Yuxiong He}{mic}
\end{icmlauthorlist}

\icmlaffiliation{mic}{Microsoft}
\icmlaffiliation{ur}{Department of Computer Science, University of Rochester}
\icmlaffiliation{eth}{Department of Computer Science, ETH Zurich}
\icmlcorrespondingauthor{Yuxiong He}{yuxhe@microsoft.com}
% \icmlkeywords{Machine Learning, ICML}
\vskip 0.3in
]

\printAffiliationsAndNotice{}

\begin{abstract}
Scalable training of large models (like BERT and GPT-3) requires careful optimization rooted in model design, architecture, and system capabilities. From a system standpoint, communication has become a major bottleneck, especially on commodity systems with standard TCP interconnects that offer limited network bandwidth. Communication compression is an important technique to reduce training time on such systems. One of the most effective methods is error-compensated compression, which offers robust convergence speed even under 1-bit compression. However, state-of-the-art error compensation techniques only work with basic optimizers like \textbf{SGD} and  \textbf{Momentum SGD}, which are linearly dependent on the gradients. They do not work with non-linear gradient-based optimizers like \textbf{Adam}, which offer state-of-the-art convergence efficiency and accuracy for models like BERT. In this paper, we propose \textbf{\OA} that reduces the communication volume by up to $5\times$, offers much better scalability, and provides the same sample-wise convergence speed as uncompressed {Adam}. Our key finding is that {Adam}'s variance (non-linear term) becomes stable
during training, hence we can run Adam in the beginning (warmup phase) and use it as a precondition for Momentum SGD during the rest of the training (compression phase). Experiments on up to 256 GPUs show that {\OA} enables up to $3.3\times$ higher throughput for BERT-Large pre-training and up to $2.9\times$ higher throughput for SQuAD fine-tuning. In addition, we provide theoretical analysis for our proposed work.
\end{abstract}

\section{Introduction}

%%%%%%%%%%%%%%%%%%%%%%%%%%%%%%%%%%
Modern advancement of machine learning is heavily driven by the advancement of computational power and techniques. Nowadays, it is not unusual to train a single model using hundreds of computational devices such as GPUs. As a result, scaling up training algorithms in the distributed setting has attracted intensive interests over the years. One important direction is communication efficient distributed training, which enhances the scalability of the training system by reducing the communication cost. Example techniques include quantization~\citep{pmlr-v70-zhang17e,Wangni2018-ux}, decentralization~\citep{Lian2017-ni, Koloskova*2020Decentralized, NIPS2018_8028}, and asynchronous communication~\citep{DBLP:journals/corr/ZhengMWCYML16, NIPS2015_6031}.

One widely used strategy for alleviating the communication overhead is gradient compression. Before communication, the original gradient $\g$ will be compressed into $\mathcal{C}_{\omega}[\g]$, where $\C_{\omega}[\cdot]$ is the compress operator{\footnote{$\C_{\omega}[\cdot]$ could also include randomness.}}. As a result the communication volume could be greatly reduced. However, this gradient compression could slow down the convergence speed because important information might get lost during the compression. To recover this information lost, error-compensated compression strategy was proposed: Instead of compressing the gradient at $t$-th iteration directly, we would first add back the compression error from the last step and then do the compression. Recent studies \citep{martinmemory}
observed that by using error-compensated compression, the asymptotic convergence speed remains unchanged for {SGD} even using 1-bit compression.

% The updating rule follows:
% \begin{align*}
% \hat{\g}_t =& \C_{\omega}[\g_t + \bdelta_{t-1}],\numberthis\label{update:ec_g}\\
% \bdelta_t = & \g_t + \bdelta_{t-1} - \hat{\g}_t,\numberthis\label{update:ec_delta}
% \end{align*}
% and we use $\hat{\g}_t$ for communication. 

On the other hand, many state-of-the-art models have to be trained using a more complicated variant, {Adam} \citep{adam}. For example, to train models such as BERT, one has to resort to the {Adam} optimizer, since training it with vanilla/momentum {SGD} has been shown to be less effective. Unfortunately, we find that error-compensated compression does not work for {Adam}, because {Adam} is non-linearly dependent on the gradient which affects the error compensation mechanism (see Section~\ref{sec:moti-convergence} and~\ref{intuition:why_adam_fails} for more details).

In this paper, we first analyze the limitation of directly applying existing compression technique to {Adam}. One of our key findings is that Adam's variance (the non-linear term) becomes stable at early stage of training (Section~\ref{sec:moti-variance}). This motivates us to design a new 2-stage algorithm, {\OA}, which uses {Adam} (warmup stage) to ``pre-condition'' a communication compressed momentum {SGD} algoirthm (compression stage). We provide theoretical analysis on communication compressed momentum {SGD}, which is the core component of {\OA}. We design a custom collective primitive using MPI to transfer the $5\times$ communication volume reduction (achieved by our algorithm) into actual runtime speedup, which is hard to accomplish using existing DL framework libraries. Experiments with BERT pre-training/fine-tuning, ResNet training and DCGAN training tasks on up to 256 GPUs show that {{\OA}} has same sample-wise convergence speed as uncompressed {Adam}, and runs up to $3.3\times$ faster than uncompressed algorithms.

{\bf (Contributions)}
We make the following contributions:
\vspace{-0.3cm}
\begin{itemize}
% %
% \item We first investigate whether we can directly
% apply standard communication compression
% techniques to the a distributed {Adam} optimizer.
% We analyze the difficulties involved in this process
% which will inspire our other technical contributions.
% %
\item We propose a new algorithm, {\OA}, a communication efficient momentum {SGD} algorithm pre-conditioned with {Adam} optimizer, which to the best of our knowledge is the first work that apply a pre-conditioned strategy for compressed momentum {SGD}. We present theoretical analysis on the convergence of {\OA}, and show that it admits the same asymptotic convergence rate as the uncompressed one.

\item We conduct experiments on large scale ML tasks that are currently challenging for {SGD} to train. We show that {\OA} is able to achieve the same convergence behaviour and final accuracy as {Adam}, together with up to $5\times$ less communication volume and $3.3\times$ faster end-to-end throughput (including the full-precision warmup stage). To the best of our knowledge, this is the first distributed learning algorithm with communication compression that can train a model as demanding as BERT.

\item We implement a custom collective communication primitive using Message Passing Interface (MPI) to provide a scalable and efficient communication system for {\OA}.

\item The {\OA} optimizer and the communication primitive backend have been open sourced in a deep learning optimization library called DeepSpeed\footnote{https://github.com/microsoft/DeepSpeed}.
\end{itemize}

\section{Related Work}
\paragraph{Communication-efficient distributed learning:}
To further reduce the communication overhead, one promising direction is to compress the variables that are sent between different workers ~\citep{NIPS2019_8694,NIPS2019_9473}. Previous work has applied a
range of techniques such as quantizaiton,
sparsification, and sketching
~\citep{Alistarh2017-yh,Agarwal2018-hg,Spring2019-ep,Ye2018-mf,MLSYS2021_Shaohuai}.
The compression is mostly assumed to be unbiased ~\citep{Wangni2018-ux,pmlr-v80-shen18a,pmlr-v70-zhang17e,NIPS2017_6749,NIPS2018_7519}.
A general theoretical analysis of centralized compressed parallel {SGD} can be found in ~\citet{Alistarh2017-yh}. Beyond this, some biased compressing methods are also proposed and proven to be quite efficient in reducing the communication cost. One example is the  \textbf{1-bit SGD} ~\citep{1-bitexp}, which compresses the entries in gradient vector into $\pm 1$ depends on its sign. The theoretical guarantee of this method is given in ~\citet{Bernstein:2018aa}.

% \vspace{-0.3cm}
\paragraph{Error-compensated compression:}
The idea of using error compensation for compression is proposed in ~\citet{1-bitexp}, where they find that by using error compensation the training could still achieves a very good speed even using $1$-bit compression. Recent study indicates that this strategy admits the same asymptotic convergence rate as the uncompressed one~\citep{martinmemory}, which means that the influence of compression is trivial. More importantly, by using error compensation, it has been proved that we can use almost any compression methods~\citep{martinmemory}, whereas naive compression could only converge when the compression is unbiased (the expectation of the compressed tensor is the same as the original). This method can be combined with decentralized training \citep{ec_decentralize}, local SGD \citep{ec_local}, accelerated algorithms \citep{ec_linearly}.  Due to the promising efficiency of this method, error compensation has been applied into many related area ~\citep{NIPS2019_9321,9051706,NIPS2019_8694,8884924,NIPS2019_9473,NIPS2019_8598,NIPS2019_9610,NIPS2019_9571} in order to reduce the communication cost. 

\paragraph{{Adam}:} {Adam}~\citep{Kingma2015AdamAM} has shown
promising speed for many deep learning tasks, and offers very good robustness against the choice of the hyper-parameters, such as learning rate. 
It can be viewed as an adaptive method that scales the learning rate with the magnitude of the gradients on each coordinate when running {SGD}. Beyond {Adam},  many other strategies that that shares the same idea of changing learning rate dynamically was studied. For example,  \citet{JMLR:v12:duchi11a} (\textbf{Adagrad}) and \citep{rmsprop} (\textbf{RMSprop}),  use the gradient, instead of momentum, for updating the parameters;  \textbf{Adadelta}~\citep{DBLP:journals/corr/abs-1212-5701} changes the variance term of {Adam} into a non-decreasing updating rule; \citet{luo2018adaptive} proposed \textbf{AdaBound} that gives both upper bound and lower bound for the variance term. In \citet{adam_theoretical,adam_liu2020adam} authors develop a novel analysis for the convergence rate of {Adam}. 

\section{Motivation and Insights}
\subsection{Communication overhead affects the efficiency of distributed training}
\label{sec:moti-profile}
To demonstrate the opportunity for communication compression, we conduct performance profiling experiments that measures the impact of communication time with respect to the total training time per step. Here, we use BERT-Large pre-training task as an example (sequence length 128, detailed training parameters can be found at Section~\ref{sec:bert-eval}), since BERT and transformer models in general are the state-of-the-art approaches in natural language processing and many other areas. We evaluate two different kinds of clusters: the first cluster has 4 NVIDIA Tesla V100 GPUs per node, and different nodes are connected by 40 Gigabit Ethernet (effective bandwidth is 4.1 Gbps based on iperf benchmark); the second cluster has 8 V100 GPUs per node, and different nodes are connected by 100 Gigabit InfiniBand EDR (effective bandwidth is close to theoretical peak based on microbenchmark). We perform BERT-Large pre-training using the two clusters with different number of nodes and GPUs, batch sizes, and gradient accumulation steps. And we measure the average latency of forward, backward (allreduce and everything else), and step function calls. Table~\ref{table_comm_overhead} presents the profiling results.

Results show that allreduce communication contributes to a great portion of the training time per step, up to 94\% and 75\% for our experiments on two above mentioned clusters with different inter-node networks. As expected, communication overhead is proportionally larger when the number of nodes is larger, when the batch size/gradient accumulation step is smaller, or when the network bandwidth is lower. These are the situations where communication compression could provide the most benefit.

\begin{table*}
  \footnotesize
  \caption{BERT-Large pre-training sequence 128 profiling results.}\label{table_comm_overhead}
  \centering
  \begin{tabular}{rrrrrrrrrrr}
  \hline
  Cluster& Num.& Num.& Batch& Batch& Grad& Forward& Backward& Backward& Step& allreduce\% \\
  Network& node& GPU& size per& size& accum.& (ms)& allreduce& everything& (ms)&  \\
  Type& & & GPU& & step& & (ms)& else (ms)& &  \\
  \hline
  Ethernet& 16& 64& 1& 64& 1& 36.65& 2205.86& 33.63& 74.96& \textbf{94\%} \\
  Ethernet& 16& 64& 16& 1024& 1& 35.71& 2275.43& 60.81& 75.59& 93\% \\
  Ethernet& 16& 64& 16& 4096& 4& 137.80& 2259.36& 243.72& 74.92& 83\% \\
  Ethernet& 8& 32& 16& 512& 1& 37.91& 2173.35& 60.71& 75.63& 93\% \\
%   Ethernet& 8& 32& 16& 2048& 4& 139.64& 2077.97& 266.10& 74.75& 81\% \\
  Ethernet& 4& 16& 16& 256& 1& 36.94& 2133.24& 62.82& 76.85& 92\% \\
%   Ethernet& 4& 16& 16& 1024& 4& 140.76& 2058.27& 246.96& 74.65& 82\% \\
  Ethernet& 2& 8& 16& 128& 1& 34.95& 1897.21& 61.23& 75.26& 92\% \\
%   Ethernet& 2& 8& 16& 512& 4& 135.64& 1817.70& 247.60& 73.92& 80\% \\
  Ethernet& 1& 4& 16& 64& 1& 35.99& 239.76& 59.95& 74.21& 58\% \\
%   Ethernet& 1& 4& 16& 256& 4& 136.16& 245.69& 252.36& 73.21& 35\% \\
  \hline
  InfiniBand& 8& 64& 1& 64& 1& 25.36& 316.18& 23.25& 58.49& \textbf{75\%} \\
  InfiniBand& 8& 64& 16& 1024& 1& 32.81& 336.40& 59.99& 57.79& 69\% \\
  InfiniBand& 8& 64& 16& 4096& 4& 131.04& 339.52& 237.92& 56.91& 44\% \\
  InfiniBand& 4& 32& 16& 512& 1& 33.45& 297.28& 56.81& 57.98& 67\% \\
%   InfiniBand& 4& 32& 16& 2048& 4& 130.96& 308.84& 239.76& 57.46& 42\% \\
  InfiniBand& 2& 16& 16& 256& 1& 32.86& 183.74& 56.49& 58.60& 55\% \\
%   InfiniBand& 2& 16& 16& 1024& 4& 131.52& 191.31& 242.87& 58.39& 31\% \\
  InfiniBand& 1& 8& 16& 128& 1& 32.74& 28.18& 59.73& 57.29& 16\% \\
%   InfiniBand& 1& 8& 16& 512& 4& 131.42& 30.96& 238.44& 57.04& 7\% \\
  \hline
  \end{tabular}\vspace{-0.5cm}
\end{table*}

\subsection{Basic compression affects {Adam}'s convergence}
\label{sec:moti-convergence}
\begin{figure}[t]
\centering
\includegraphics[width=0.45\textwidth]{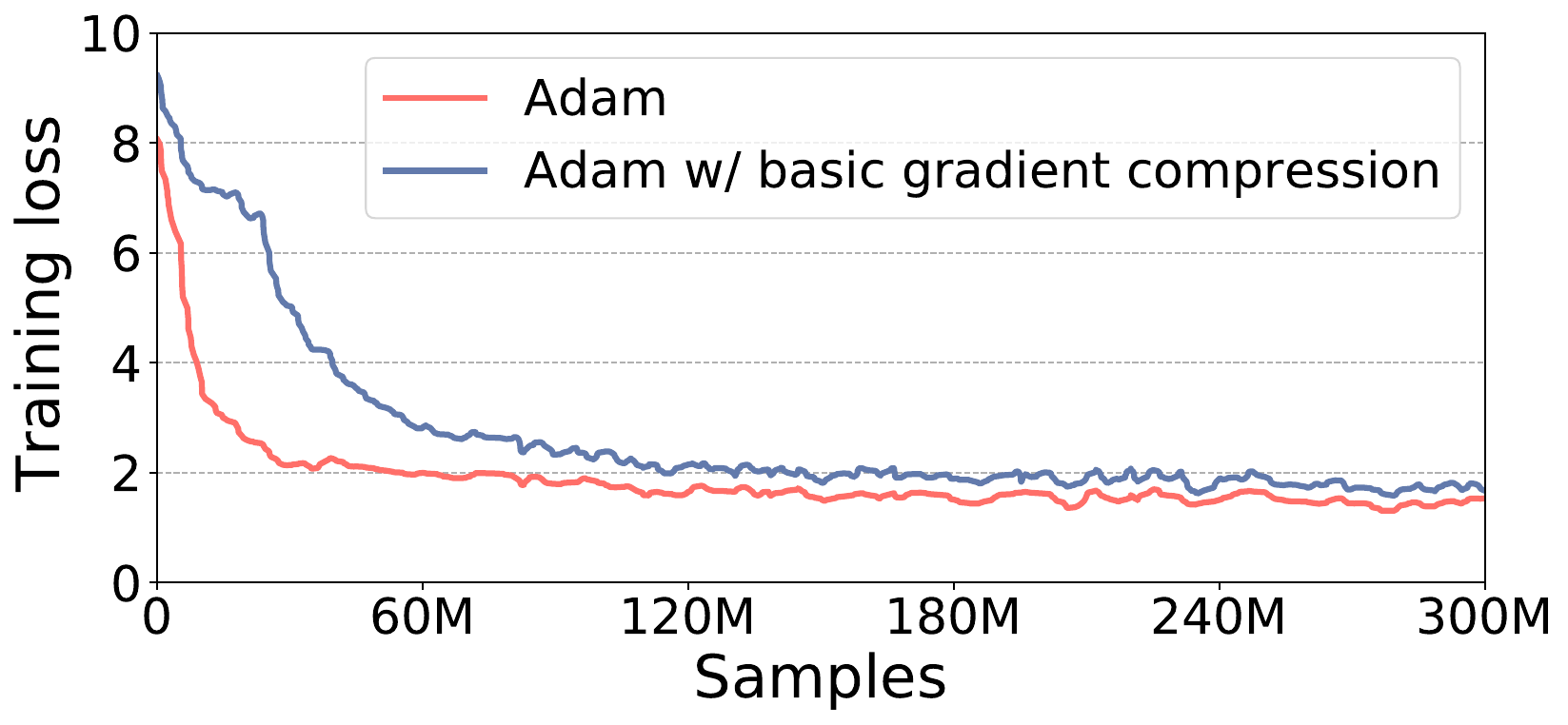}
\caption{Training loss for BERT-Large pre-training using vanilla Adam and Adam with error compensated gradient compression.}\label{fig:moti_loss}\vspace{-0.4cm}
\end{figure}

Given the great opportunity for communication compression, we investigate whether existing error-compensated gradient compression strategy can be applied to {Adam}, an important optimization algorithm for large model distributed training. We implement a basic compression strategy for {Adam} based on the compression-based {SGD} approach~\citep{martinmemory}, where we perform error-compensated 1-bit compression over the gradient, and update both the momentum and variance based on the compressed gradient. We compare the BERT-Large pre-training (sequence length 128) training loss when using vanilla {Adam} and {Adam} with our basic compression strategy in Figure~\ref{fig:moti_loss}.

Results show that basic compression based on existing work greatly affects the convergence speed for Adam. The main reason is that {Adam} is non-linearly dependent to the gradients (see Section~\ref{intuition:why_adam_fails} for more details).  This motivates us to look for novel compression strategy that can overcome the non-linear gradient dependency challenge, and also achieve the same convergence speed as {Adam}.

\subsection{{Adam}'s variance becomes stable during training}
\label{sec:moti-variance}
Unlike {SGD}, which  directly uses the gradient $\g$ to update the model $\x$, {Adam} uses two auxiliary variables $\m$ and $\v$ for the update. The mathematical updating rule of  original {Adam} can be summarized as:
\begin{align*}
\bm{m}_{t+1} =& \beta_1\bm{m}_t + (1-\beta_1)\g_t\\
\bm{v}_{t+1} =&  \beta_2\bm{v}_t + (1-\beta_2)(\bm{g}_t)^2,\numberthis\label{alg:v}\\
\bm{\x}_{t+1} =&  \x_t - \gamma\frac{\m_{t+1}}{\sqrt{\v_{t+1} }+ \eta}
\end{align*}
Here $\x_t$ is the model at $t$-iteration, $\g_t = \nabla F(\x_t;\bzeta_t)$ is the stochastic gradient, $\gamma$ is the learning rate, $\eta$ usually is a very small constant, $\beta_1$ and $\beta_2$ are decaying factor that controls the speed of forgetting history information. Notice that we disable the bias correction term in the original {Adam}, which is consistent with exact optimizer for training BERT \citep{bert}.

Here we refer $\m_t$ as the momentum term and $\v_t$ as the variance term. Notice that when $\v_t$ is changed into a constant $\v$, then {Adam} becomes equivalent to {Momentum SGD} under a coordinate-dependent learning rate $\frac{\gamma}{\sqrt{\v} + \eta}$.

\begin{figure}[t]
\centering
\includegraphics[width=0.45\textwidth]{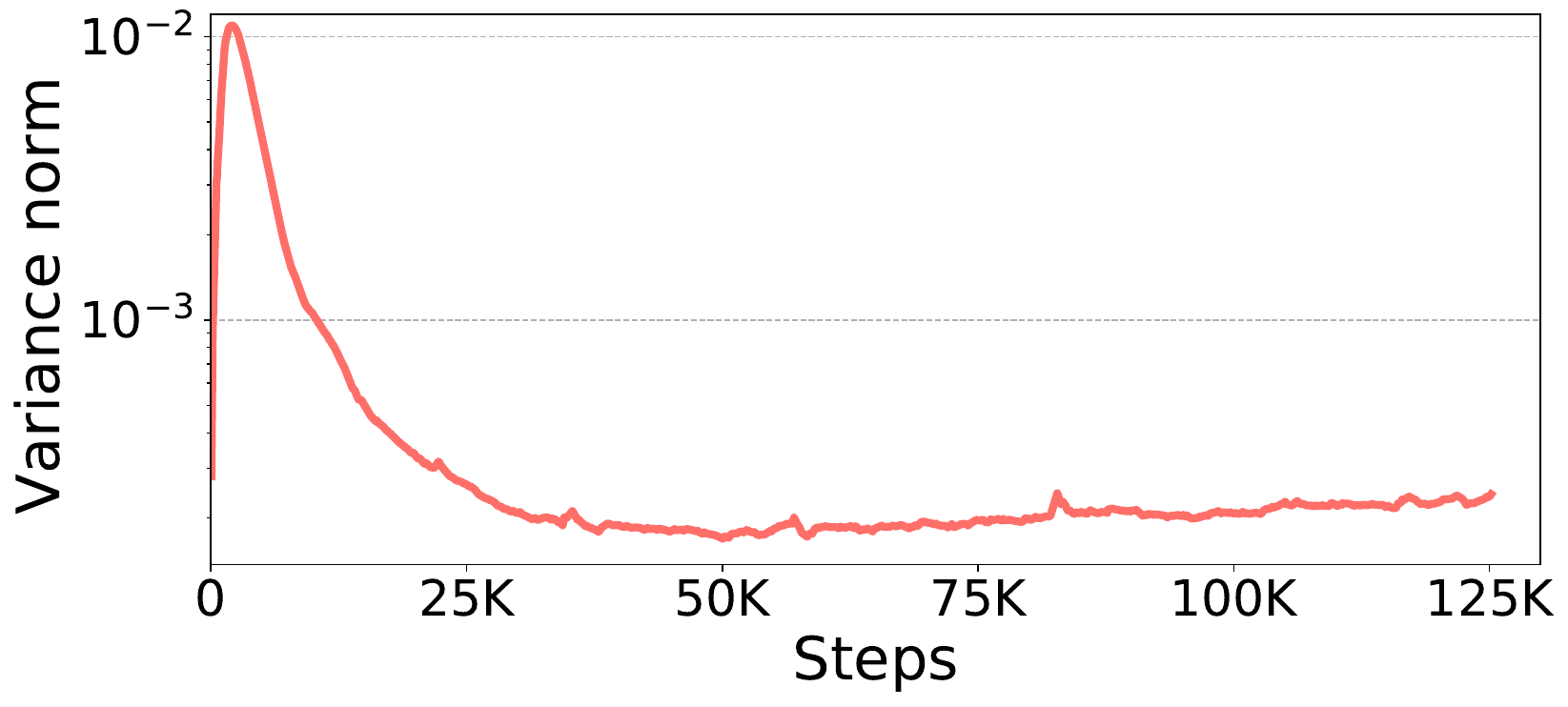}
\caption{Norm of fused variance for BERT-Large pre-training using vanilla Adam. The y-axis is in log scale.}\label{fig:moti_var_norm_log}\vspace{-0.5cm}
\end{figure}

To investigate the non-linear gradient dependency challenge, we analyze {Adam}'s variance during BERT-Large pre-training (sequence length 128). At each step, we fuse the variance of all parameters, and calculate the norm of the fused variance. Figure~\ref{fig:moti_var_norm_log} presents this fused variance norm at each step. Results show that the variance norm becomes stable after around $23K$ steps. This motivates our approach {\OA} to ``freeze'' the Adam variance after it becomes stable, and then use it as a precondition during 1-bit compression stage.

\section{{\OA} Algorithm}
In this section, we start with some background introduction for error compensated compression and why it is incompatible with {Adam}. Then we give full description of {\OA}.

\paragraph*{Problem setting} In this paper, we focus on the following optimization task and rely on the following notions and definitions:
\vspace{-0.3cm}
\begin{equation}
\min_{\bm{x}\in\mathcal{R}^d}\quad f(\bm{x}) = {\frac{1}{n}} \sum_{i=1}^n \underbrace{\mathbb{E}_{\bm{\zeta}^{(i)}\sim\mathcal{D}_i}F(\bm{x}; \bm{\bm{\zeta}}^{(i)})}_{:=f_i(\x)},\label{eq:main}
\end{equation}
where $d$ is the dimension of the input model $\x$, $n$ is the number of workers included, $\mathcal{D}_i$ is the data distribution of individual data sample $\bzeta^\ti$ on the $i$-th worker, $F(\x;\bzeta)$ is the loss function.

\paragraph{Notations and definitions}
Throughout this paper, we use the following notations:
\begin{itemize}
\item $\nabla f(\cdot)$ denotes the gradient of a function $f$.
\item $f^{*}$ denotes the optimal value of the minimization problem \eqref{eq:main}.
\item $f_i(\x) := \mathbb{E}_{\bzeta^\ti\sim\mathcal{D}_i}F(\x; \bzeta^\ti)$.
\item $\|\cdot\|$ denotes the $\ell_2$ norm for vectors and the spectral norm for matrices.
\item $\|X\|_A:=\text{Tr}(X^{\top}AX)$.
\item $\bm{C}_{\omega}(\cdot)$ denotes the randomized compressing operator.
% where $\omega$ denotes the random variable. One example is the randomized quantization operator, for example, $\bm{C}_{\omega}(0.7) = 1$ with probability $0.7$ and $\bm{C}_{\omega}(0.7) = 0$ with probability $0.3$. 
% It is also worth noting that this notion $\bm{C}_{\omega}(\cdot)$ also covers the deterministic scenario, for example, $\bm{C}_{\omega}(\cdot)$ is a one bit compression operator.
\item { $\sqrt{\cdot}$ denotes the square root of the argument. In this paper if the argument is a vector, then it returns a vector taking the element-wise square root.}
\item $(\x)^2$ denotes the element-wise square operation if $\x$ is a vector.
\item $\frac{\bm{a}}{\bm{b}}$ or $\bm{a}/\bm{b}$ denotes the element-wise division operation if both $\bm{a}$ and $\bm{b}$ are vectors and their dimension matches.
% \item { $\oslash$ denotes the element-wise divide operator, that is, the $i$th element of $\bm{m} \oslash \bm{v}$ is $\bm{m}_i / \bm{v}_i$.}
% \item $\odot$ denotes the element-wise multiply operator, that is, the $i$th element of $\bm{m} \odot \bm{v}$ is $\bm{m}_i * \bm{v}_i$.
\end{itemize}

\subsection{Why error compensation works for {SGD}}
For  {SGD} , since the update is linearly dependent to the gradient, using error compensation could potentially remove the side-effect of the history compression error. The updating rule of \textbf{vanilla SGD} follows
\begin{align*}
\x_{t+1} =& \x_t - \gamma \g_t  = \x_0 - \gamma\sum_{s=0}^t\g_s.\numberthis\label{intuition:sgd_eq1}
\end{align*}
When directly compressing the gradient without error compensation, the updating rule becomes
\begin{align*}
\x_{t+1} =& \x_t - \gamma C_\omega[\g_t] =  \x_t - \gamma (\g_t-\bdelta_t)\\
= &\x_0 - \gamma\sum_{s=0}^t\g_s + \underbrace{\gamma\sum_{s=0}^t \bdelta_s}_{\text{history compression error}}.\numberthis\label{intuition:sgd_eq2}
\end{align*}
As we can see in \eqref{intuition:sgd_eq2}, the history compression error would get accumulated and therefore slow down the convergence rate. Moreover, previous work \citep{Alistarh2017-yh} indicates that when using biased compression operator, the training convergence cannot be guaranteed. 

Now if we apply error compensation at each compression step, the updating rule becomes
\begin{align*}
\x_{t+1} =& \x_t - \gamma C_\omega[\g_t + \bdelta_{t-1}] =  \x_t - \gamma (\g_t-\underbrace{\bdelta_t +\bdelta_{t-1}}_{\text{error cancellation}})\\
=& \x_0 - \gamma\sum_{s=0}^t\g_s + \gamma\sum_{s=0}^t(\bdelta_s - \bdelta_{s-1})\\
=& \x_0 - \gamma\sum_{s=0}^t\g_s + \gamma\bdelta_t.\numberthis\label{intuition:sgd_eq3}
\end{align*}

This demonstrates that by using error compensation, each step's compression error would get cancelled in the next step instead of getting accumulated over steps. To make the error compensation work correctly, it is necessary that we ensure an error cancellation term $\bdelta_t +\bdelta_{t-1}$ in the updating rule. Below we are going to see that this cannot be achieved for {Adam}.

\subsection{Why {Adam} cannot be combined with error compensation}\label{intuition:why_adam_fails}
As we can see, {Adam} is non-linearly dependent to the gradient, and this non-linearity is widely believed to be essential for the superiority of {Adam}. Here we discuss why this non-linearity makes {Adam} incompatible with error compensation.

\paragraph{Difficulty for estimating the variance term $\v$.} Notice that for {Adam}, it is necessary to communicate the gradient $\g_t$ or momentum $\m_t$, and the variance term can be updated using $\g_t$. However, when using error-compensated gradient to update $\v_t$, the updating rule follows:
\begin{align*}
\v_{t+1} = &\beta_2 \v_t + (1-\beta_2)\left(C_\omega[\g_t + \bdelta_{t-1}]  \right)^2\\
=& \beta_2 \v_t + (1-\beta_2)\left(\g_t + \bdelta_{t-1} - \bdelta_t  \right)^2\\
= & \beta_2 \v_t + (1-\beta_2)\left(\g_t   \right)^2  +   \underbrace{\left( \bdelta_{t-1} - \bdelta_t  \right)^2}_{\text{non-linear error correction}}\\
& + 2 \langle \g_t,\bdelta_{t-1} - \bdelta_t\rangle.
\end{align*}
Here the quadratic term $\left( \bdelta_{t-1} - \bdelta_t  \right)^2$ cannot be cancelled by itself, therefore it is hard to get an accurate estimation of $\v_t$ since the history error do not cancel.
\paragraph{Difficulty for setting the correction factor.} Another problem is that for  {SGD} , when applying error compensation under a time varying learning rate $\gamma_t$, we need to compensate the history error using
\begin{align*}
C\left[ \g_t + \frac{\gamma_t}{\gamma_{t-1}}\bdelta_{t-1}\right],
\end{align*} instead of adding back $\bdelta_{t-1}$ directly. In this case,
if we view $\frac{\gamma}{\sqrt{\v_t} + \eta}$ as a coordinate-dependent learning rate, which makes {Adam} equivalent to  {Momentum SGD} with time-varying learning rate, we need to apply the scale factor according to 
\begin{align*}
\m_{t+1} = C_\omega\left[\beta_1\m_t + (1-\beta_1)\g_t + \frac{\sqrt{\v_{t-1}} + \eta}{\sqrt{\v_{t}} + \eta}\bdelta_{t-1}\right].
\end{align*}
The problem is that we cannot get the value of $\v_{t}$ after the compression, which makes it impossible to set the scale factor for error compensation.

% This leads us to {\OA}, which is capable of achieving almost the same convergence rate and can be easily combined with error compensation. 

\begin{figure*}[t]
\centering
\subfigure[\scriptsize \textbf{All-to-All step}: Each worker sends its $i$-th chunk to worker $i$.]{
\begin{minipage}[t]{0.3\linewidth}
\centering
\includegraphics[width=1\textwidth]{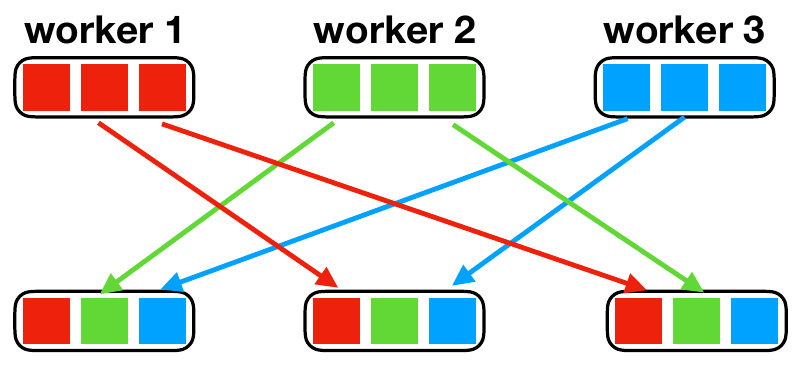}
\end{minipage}
}\quad
\subfigure[\scriptsize \textbf{Average step}: Each worker averages all chunks it receives.]{
\begin{minipage}[t]{0.3\linewidth}
\centering
\includegraphics[width=1\textwidth]{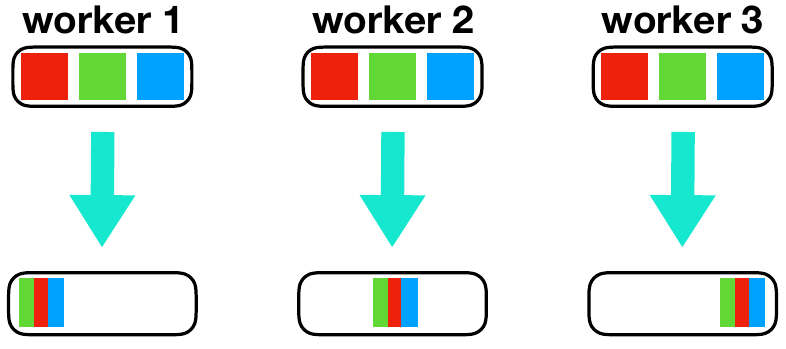}
\end{minipage}%
}\quad
\subfigure[\scriptsize \textbf{All-Gather step}: Each worker receives the $i$-th chunk from worker $i$.]{
\begin{minipage}[t]{0.3\linewidth}
\centering
\includegraphics[width=1\textwidth]{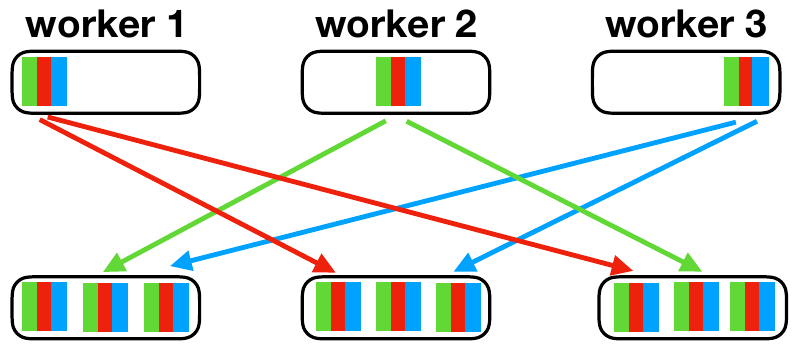}
\end{minipage}%
}%
\centering
\caption{Efficient system design for communication (compressed\_allreduce)}\label{allreduce}\vspace{-0.5cm}
\label{fig:allreduce}
\end{figure*}

\subsection{{\OA}}\label{alg:description}
Based on our findings (Section~\ref{sec:moti-variance}) that {Adam}'s variance term becomes stable at an early stage, we propose {\OA} summarized in Algorithm \ref{alg:de_ec}. First we use vanilla {Adam} for a few epochs as a warm-up. After the warm-up stage, the compression stage starts and we stop updating the variance term $\v$ and use it as a fixed precondition. At the compression stage, we communicate based on the momentum applied with error-compensated 1-bit compression. The momentums are quantized into 1-bit representation (the sign of each element). Accompanying the vector, a scaling factor is computed as $\frac{\text{magnitude of compensated gradient}}{\text{magnitude of quantized gradient}}.$ This scaling factor ensures that the compressed momentum has the same magnitude as the uncompressed momentum. This 1-bit compression could reduce the communication cost by $97\%$ and $94\%$ compared to the original float32 and float16 training, respectively.

\begin{algorithm}[t!]\caption{{\OA}}
\begin{algorithmic}[1]
\footnotesize
\STATE {\bfseries Initialize}: $\x_0$, learning rate $\gamma$,  initial error $\bdelta = \boldsymbol{0}$, $\m_0 = \boldsymbol{0}$, $\v_0 = \boldsymbol{0}$, number of total iterations $T$,  warm-up steps $T_{w}$, two decaying factor $\beta_1$, $\beta_2$ and $\eta$ for {Adam}.

\STATE Running the original {Adam} for $T_{w}$ steps, then store the variance term (defined as $\v_t$ in \eqref{alg:v}) $\bm{v}_{_{ T_w}}$.
\FOR {$t=T_w,\ldots,T$}

\STATE \textbf{(On $i$-th node)}
\STATE  Randomly sample $\bm{\zeta}_t^{(i)}$  and compute local stochastic gradient $\g_t^{(i)} := \nabla F_i(\x_t^{(i)}, \bm{\zeta}_t^{(i)})$.

\STATE Update the local momentum variable $\m_{t-1}$ according to
$
\m_t^{(i)} =  \beta_1\bm{m}_{t-1} + (1 - \beta_1)\g_t^{(i)}.
$
\STATE  Compress $\m_t^{(i)}$ into $\hat{\m}_t^\ti = \bm{C}_\omega\left[\m_t^{(i)} + \bdelta_{t-1}^{(i)}\right]$, and update the compression error by $\bdelta_t^{(i)} = \m_t^{(i)} + \bdelta_{t-1}^{(i)} - \hat{\m}_t^\ti$.
\STATE Send the  $\hat{\m}_t^\ti$ to the server.
\STATE \textbf{(On server)}
\STATE Take the average over all $\hat{\m}_t^\ti$ it receives and compress it into
$
\overline{\m}_t =\bm{C}_\omega\left[ \frac{1}{n}\sum_{i=1}^n\hat{\m}_t^\ti + \overline{\bdelta}_{t-1}\right],
$
 and update the compression error accordingly by $\overline{\bdelta}_t = \frac{1}{n}\sum_{i=1}^n\hat{\m}_t^\ti + \overline{\bdelta}_{t-1} - \overline{\m}_t$.
 \STATE Send $\overline{\m}_t$ to all the workers.
 \STATE \textbf{(On $i$-th node)}
 \STATE Set $\m_t = \overline{\m}_t$ , and update local model $\x_{t+1} = \x_t - \gamma \m_t/\sqrt{\bm{v}_{_{\tiny T_w}}} $.
\ENDFOR
\STATE {\bfseries Output}: $\x$.
\end{algorithmic}\label{alg:de_ec}
\end{algorithm}

\section{Theoretical Analysis}
Notice that for {\OA}, we only use original {Adam} at warm-up, and then we essentially run error-compensated momentum {SGD} with coordinate-dependent learning rate $\frac{\gamma}{\sqrt{\v_{_{T_w}}}}$. Therefore here we consider the {Adam}-based warm-up phase as a way to find a good precondition variance term $\v_{_{T_w}}$ to be used in the compression phase. Below we are going to introduce the convergence rate for the compression phase after warm-up. We first introduce some necessary assumptions, then we present the theoretical guarantee of the convergence rate for {\OA}.

\begin{assumption}\label{ass:global}
We make the following assumptions:
\begin{enumerate}
\item \textbf{Lipschitzian gradient:} $f(\cdot)$ is assumed to be  with $L$-Lipschitzian gradients, which means
  \begin{align*}
  \|\nabla f(\bm{x}) - \nabla f(\bm{y}) \| \leq L \|\bm{x} - \bm{y} \|,\quad \forall \bm{x},\forall \bm{y},
  \end{align*}
 \item\label{ass:var} \textbf{Bounded variance:}
The variance of the stochastic gradient is bounded
\begin{align*}
\mathbb E_{\bzeta^\ti\sim\mathcal{D}_i}\|\nabla F(\bm{x};\bm{\zeta}^\ti) - \nabla f(\bm{x})\|^2 \leq \sigma^2,\quad\forall \bm{x},\forall i.
\end{align*}
\item \textbf{Bounded magnitude  of error for $\C_{\omega}[\cdot]$:}
The magnitude of worker's local errors $\bm{\delta}_t^{(i)}$  and the server's global error $\overline{\bdelta}_t$, are assumed to be bounded by a constant $\epsilon$
\begin{align*}
\sum_{k=1}^n\mathbb E_{\omega} \left\|\bm{\delta}_t^{(i)}\right\|\leq \frac{\epsilon}{2},\quad
\sum_{i=1}^n\mathbb E_{\omega}\left\|\overline{\bdelta}_t\right\|\leq  \frac{\epsilon}{2},\quad\forall t,\forall i.
\end{align*}
\end{enumerate}
\end{assumption}
% Notice that previous work ~\citep{pmlr-v80-wu18d} needs an extra assumption to upper bound the magnitude of the gradient as $G$, which we did not require in our theoretical analysis. Therefore we need to design new proving strategies for the proof.

Next we present the main theorem for {\OA}.
\begin{theorem}\label{theo:global}
 Under Assumption~\ref{ass:global}, for {\OA}, we have the following convergence rate
 \begin{align*}
   &\left(1-\frac{\gamma L}{v_{\min}} - \frac{2\gamma^2 L^2}{(1-\beta)^2v_{\min}^2} \right)\sum_{t=0}^T \mathbb E\|\nabla f(\bm{x}_t)\|^2_{V}\\
    \leq & \frac{2\mathbb E f(\bm{x}_{0}) - 2f(\bm{x}^*)}{\gamma}   + \frac{6\gamma^2L^2\epsilon^2 T}{(1-\beta)^2v_{\min}^3}  + \\ & \frac{L\gamma \sigma^2T}{nv_{\min}} + \frac{2\gamma^2L^2\sigma^2  T}{n(1-\beta)^2v_{\min}^2},\numberthis\label{main:theo:eq}
\end{align*}
where $V= \text{diag}\left(1/\v_{T_w}^{(1)},1/\v_{T_w}^{(2)},\cdots,1/\v_{T_w}^{(d)}\right)$ is a diagonal matrix spanned by $\v_{_{T_w}}$ and $v_{\min} = \min\{\v_{T_w}^{(1)},\v_{T_w}^{(2)},\cdots,\v_{T_w}^{(d)}\}$ is the mimimum value in $\v_{T_w}$
\end{theorem}

Given the generic result in Theorem~\ref{theo:global}, we obtain the convergence rate for {\OA} with appropriately chosen learning rate $\gamma$.

\begin{corollary}\label{coro:global}
Under Assumption~\ref{ass:global}, for {\OA}, choosing
$
\gamma = \frac{1}{4L(v_{\min})^{-1} + \sigma\sqrt{\frac{ T}{n}} + \epsilon^{\frac{2}{3}} T^{\frac{1}{3}}(v_{\min})^{-1} },
$
we have the following convergence rate
\begin{align*}
\frac{1}{Tv_{\min}}\sum_{t=0}^{T-1}\mathbb{E}\|\nabla f(\bm{x}_t)\|^2_V \lesssim \frac{\sigma}{\sqrt{nT}} + \frac{\epsilon^{\frac{2}{3}}}{T^{\frac{2}{3}}} + \frac{1}{ T},
\end{align*}
where we treat $f(\bm{x}_1) - f^*$, $\beta$ and $L$ as constants.
\end{corollary}

% This result suggests that
% \begin{itemize}
% \item ({\bf Comparison to SGD}) {\OA} essentially admits the same convergence rate as {SGD} in the sense that both of them admit the asymptotical convergence rate $O(1/\sqrt{T})$;
% \item ({\bf Linear Speedup}) The asymptotical convergence rate of {\OA} is $O(1/\sqrt{nT})$, the same convergence rate as Parallel SGD. 
% \end{itemize}

This result suggests that: {\OA} essentially admits the same convergence rate as distributed {SGD} in the sense that both of them admit the asymptotical convergence rate $O(1/\sqrt{nT})$, which means we can still achieve linear speedup w.r.t. the number of workers $n$.

% ; The asymptotical convergence rate of {\OA} is $O(1/\sqrt{nT})$, the same convergence rate as Parallel SGD.

\section{Efficient system design for compressed communication}
NVIDIA NCCL is an efficient and widely used communication library that has been tightly integrated in DL frameworks like PyTorch and TensorFlow. However, NCCL library cannot be used directly for performing communication based on 1-bit compression. This is because the collective communication primitives like Allreduce and Allgather are at a higher level of abstraction and can only perform data movement and/or simple operations like sum, min, max etc. In addition, NCCL library (before v2.7) did not expose either an Alltoall primitive or any point-to-point (send/recv) communication primitives that can be used to implement an Alltoall. Thus for {\OA}, we designed a custom collective primitive using Message Passing Interface (MPI). We call it ``compressed allreduce'' and it has three phases as shown in Figure~\ref{fig:allreduce}: 1) The all-to-all step, which we have implemented using the MPI\_Alltoall (personalized exchange) primitive, 2) The average step, where {\OA} computes the average of compressed local momentums, and 3) The all-gather step, which we implement using MPI\_Allgather. We develop two versions of compressed allreduce: 1) CUDA-Aware version that exploits GPUDirect features and requires CUDA-Aware libraries like MVAPICH2-GDR and 2) Basic version that can be used with any MPI library but copies data between GPU and CPU buffers. The CUDA-Aware version works only on systems with InfiniBand whereas the basic version can run on any system with Ethernet interconnect. 

\section{Experiments}

% \begin{figure}[h]
% \centering
% \subfigure[]{
% \begin{minipage}[t]{0.45\linewidth}
% \centering
% \includegraphics[width=0.8\textwidth]{base128.pdf}
% %\caption{fig1}
% \end{minipage}%
% }%
% \subfigure[]{
% \begin{minipage}[t]{0.45\linewidth}
% \centering
% \includegraphics[width=0.8\textwidth]{large128.pdf}
% %\caption{fig2}
% \end{minipage}%
% }%
% \centering
% \caption{ pics}
% \end{figure}

We evaluate {{\OA}} and existing approaches on various training tasks such as pre-training BERT-Base and BERT-Large, fine-tuning BERT on SQuAD 1.1 and GLUE, training ResNet on CIFAR-10 and ImageNet, and training DCGAN on CelebA. We show that {{\OA}} provides same sample-wise convergence speed as uncompressed {Adam}, and runs up to 3.3 times faster than uncompressed algorithms under limited bandwidth.

%%% Conglong: I commented this section since we no longer evaluate top-k in the main paper (maybe add it to the supplementary). I moved the 1-bit compression description to section 4.3
% \subsection{Compression Method}
% For BERT-Base, BERT-Large and SQuAD 1.1, we use $1$-bit compression. For ResNet-18, we use both $1$-bit compression and Top-$k$ compression.
  
% \paragraph{1-bit compression} The gradients are quantized into 1-bit representation (the sign of each element). Accompanying the vector, a scaling factor is computed as $\frac{\text{magnitude of compensated gradient}}{\text{magnitude of quantized gradient}}.$ The scaling factor is multiplied onto the quantized gradient whenever the quantized gradient is used, so that the recovered gradient has the same magnitude of the compensated gradient. This compression could reduce the $97\%$ communication cost of the original for float32 type training and $94\%$ for float16 type training.
  
% \paragraph{Top-$k$ compression} We take top $k\%$ elements of the original gradient that is sorted by its absolute magnitude. The communication cost is reduced into $k\%$ of the original.

\subsection{BERT pre-training and fine-tuning}
\label{sec:bert-eval}
\paragraph{Dataset and models} We evaluate the convergence and performance of {\OA} and uncompressed {Adam} for BERT-Base ($L=12$, $H=768$, $A=12$, $110M$ params) and BERT-Large ($L=24$, $H=1024$, $A=16$, $340M$ params) pre-training tasks. We use the same dataset as \citet{bert}, which is a concatenation of Wikipedia and BooksCorpus with $2.5B$ and $800M$ words respectively. We use the GLUE fine-tuning benchmark\citep{glue} to evaluate the convergence of the BERT models trained by {Adam} and {\OA}.

In addition, we also evaluate the convergence and performance of {\OA} for SQuAD 1.1 fine-tuning task\footnote{https://rajpurkar.github.io/SQuAD-explorer/} using a pre-trained BERT model checkpoint from HuggingFace\footnote{https://github.com/huggingface/transformers}.

\paragraph{Hardware} For all experiments in this Section~\ref{sec:bert-eval} we use the two clusters described in Section~\ref{sec:moti-profile}. We use up to 256 GPUs for pre-training tasks and up to 32 GPUs for fine-tuning tasks.

\paragraph{Training parameters} For BERT pre-training, the learning rate linearly increases to $4\times 10^{-4}$ as a warmup in the first $12.5K$ steps, then decays into $0.99$ of the original after every $520$ steps. We set the two parameters in Algorithm~\ref{alg:de_ec} as $\beta_1 = 0.9$ and $\beta_2 = 0.999$ for {\OA} and {Adam}. For convergence test, we set total batch size as $4K$ for BERT-Base and BERT-Large. For performance test, we test different batch sizes. Table~\ref{table_bert_steps} summarizes the total number of steps for BERT sequence length 128 and 512 phases, together with the number of warmup steps for {\OA}. We manually tuned the number of warmup steps for {\OA} evaluations. On the other hand, we find that this configuration can be auto-tuned: First, the number of {\OA} warmup steps should be no less than the number of learning rate warmup steps, since Adam's variance term is unstable during LR warmup. Second, we find that the ratio $\frac{\| \v_t \|_1}{\| \v_{t-\Delta} \|_1}$ (where 
$\|\cdot\|_1$ is the  $l_1$ norm of the vector and we set $\Delta=\frac{1}{1-\beta_2}$) is a good indicator of how stable the variance term is. For BERT-Large pre-training seqlen 128, when we set a threshold of $\geq 0.96$ for this ratio, the warmup will stop at step 22173, which is very close to our manualy tuned $23K$ warmup steps.

For GLUE benchmarks we use original {Adam} optimizer and perform single-task training on the dev set. We search over the hyperparameter space with batch sizes $\in\{8,16\}$ and learning rates $\in\{1\times 10^{-5},3\times 10^{-5},5\times 10^{-5},8\times 10^{-5}\}$. Other setting are the same as pre-training task.

For SQuAD fine-tuning we use the same parameters as published by HuggingFace (batch size = $24$, learning rate=$3e-5$, dropout=$0.1$, 2 epochs), except that we increase the batch size to $96$ (using $32$ GPUs). The first $400$ steps out of total $1848$ steps are used as the warmup stage for {\OA}.

\begin{table}[t]
  \footnotesize
  \caption{Number of steps for BERT pre-training tasks.}\label{table_bert_steps}
  \centering
  \begin{tabular}{lll}
  \hline
  & Seqlen 128& Seqlen 512\\
  & (warmup)& (warmup)\\
  \hline
  BERT-Base {Adam}& $118K$ (N/A)& $22K$ (N/A) \\
  BERT-Base {\OA}& $118K$ ($16K$)& $22K$ ($1.5K$) \\
  BERT-Large {Adam}& $152K$ (N/A)& $10K$ (N/A) \\
  BERT-Large {\OA}& $152K$ ($23K$)& $10K$ ($1.5K$) \\
  \hline
  \end{tabular}\vspace{-0.5cm}
\end{table}

\begin{table*}[t]
\footnotesize
  \caption{GLUE development set results. BERT-Base/Large(original) results are from \citet{bert}. BERT-Base/Large (uncompressed) results use the full-precision \textbf{BertAdam} with the same training parameters as the {\OA} case. BERT-Base/Large (compressed) are the results using {\OA}. The scores are the median scores over 10 runs.}\label{table1}
  \centering
  \begin{tabular}{lccccccc}
  \hline  %添加表格头部粗线
  \textbf{Model}& RTE& MRPC& CoLA & SST-2& QNLI& QQP& MNLI-(m/mm) \\
  \hline  %添加表格中横线
  BERT-Base (original) & 66.4 & 84.8 & 52.1  & 93.5 & 90.5& 89.2& 84.6/83.4\\
  BERT-Base (uncompressed) & 68.2 & 84.8 & 56.8  & 91.8 & 90.9& 90.9& 83.6/83.5\\
  BERT-Base (compressed) & 69.0& 84.8 & 55.6   & 91.6 & 90.8& 90.9& 83.6/83.9\\
  \hline
  BERT-Large (original) & 70.1& 85.4 & 60.5   & 94.9 & 92.7& 89.3& 86.7/85.9\\
  BERT-Large (uncompressed) & 70.3& 86.0 & 60.3   & 93.1 & 92.2& 91.4& 86.1/86.2\\
  BERT-Large (compressed) & 70.4& 86.1 & 62.0  & 93.8 & 91.9& 91.5& 85.7/85.4\\
  \hline %添加表格底部粗线
  \end{tabular}\vspace{-0.5cm}
\end{table*}

\begin{figure}[t]
    \centering
    \subfigure[Sample-wise]{
        \centering
        \includegraphics[width=0.8\columnwidth]{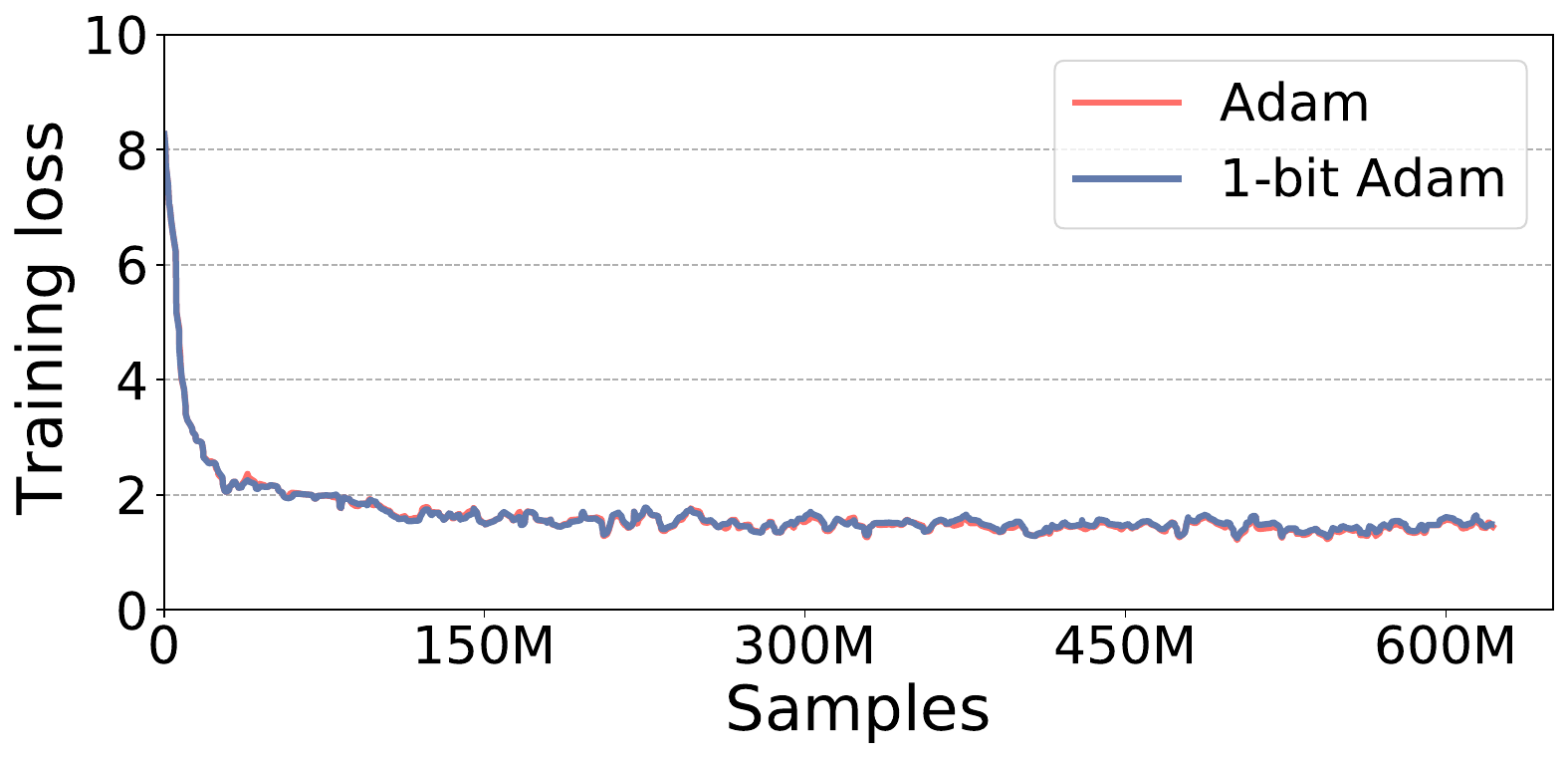}
        \label{fig:bert}
    }
    \subfigure[Time-wise]{
        \centering
        \includegraphics[width=0.8\columnwidth]{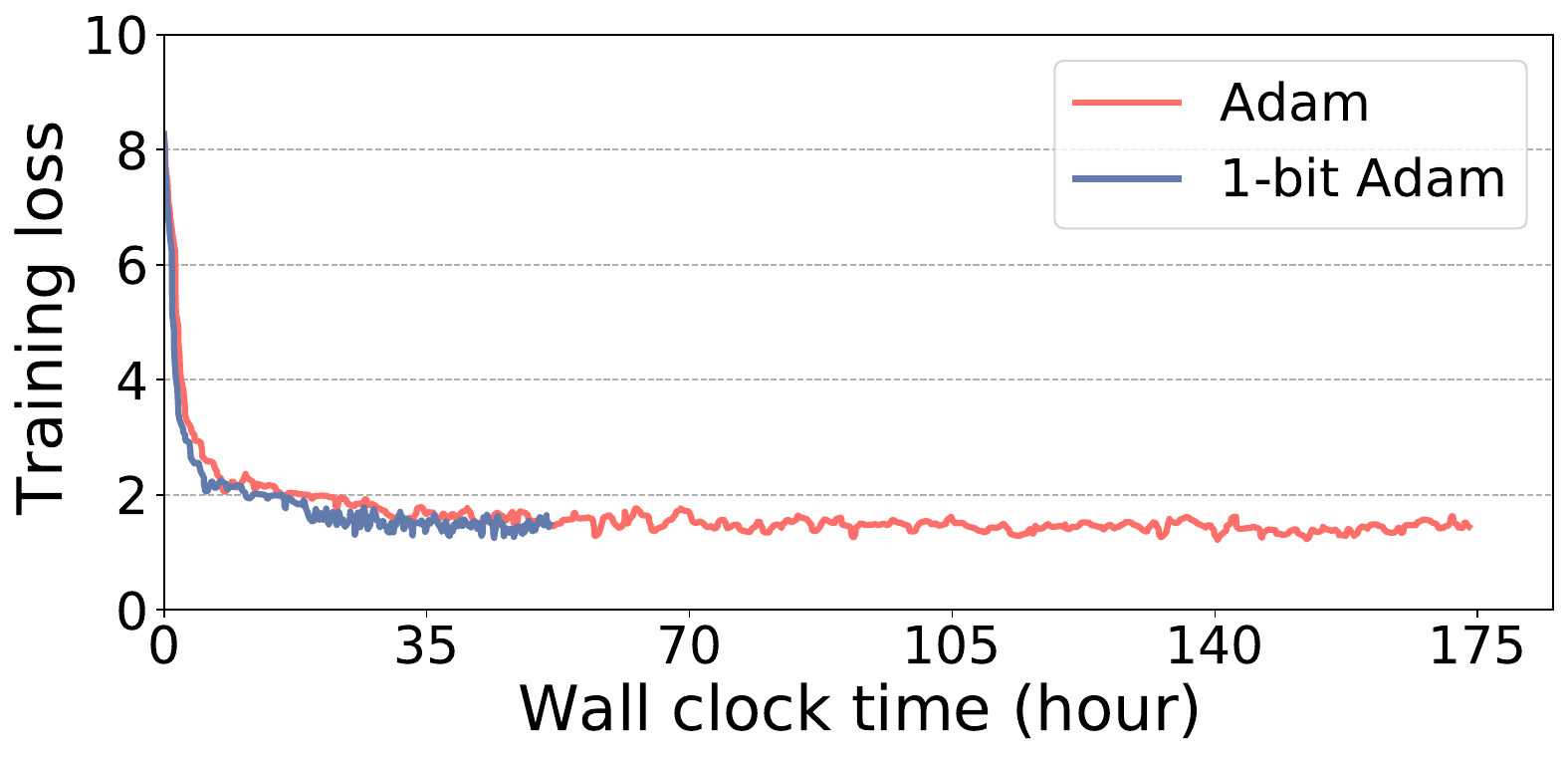}
        \label{fig:bert-time}
    }
    \caption{Sample-wise and time-wise convergence speed for BERT-Large pre-training sequence length 128 using 64 GPUs on the Ethernet cluster. {\OA} and {Adam} also achieve the same sample-convergence speed for BERT-Base pre-training.}
    \vspace*{-0.5cm}
\end{figure}

\paragraph{Convergence results}
Figure~\ref{fig:bert} presents the sample-wise convergence results. We use the BertAdam \citep{bert} optimizer as the uncompressed baseline. For both BERT-Base and BERT-Large and for both sequence length phases, we find that {\OA} provides the same convergence speed as baseline, while the communication volume is reduced into $6\%$ of the original during the compression stage.

Table~\ref{table1} presents the GLUE results using the checkpoints from our pre-training experiments. {\OA} achieves similar accuracy compared to the uncompressed baseline and the numbers reported in previous work.

For SQuAD 1.1 fine-tuning task using checkpoint from  HuggingFace, {\OA} achieves similar F1 score (93.32) compared to the score reported by HuggingFace (93.33) using same number of samples and trainig parameters. 
  
% \begin{figure}[t]
% \centering
% \includegraphics[width=0.45\textwidth]{large_loss.pdf}
% \caption{Epoch-wise convergence speed for BERT-Large pre-training sequence length 128. {\OA} and {Adam} also achieve the same convergence speed for BERT-Base pre-training.}\label{fig:bert}\vspace{-0.5cm}
% \end{figure}

\begin{figure*}[t]
\centering
\subfigure[Bert-Large pre-training, batch size = number of GPUs $\times$ 16]{
\begin{minipage}[t]{0.34\linewidth}
\centering
\includegraphics[width=1\textwidth]{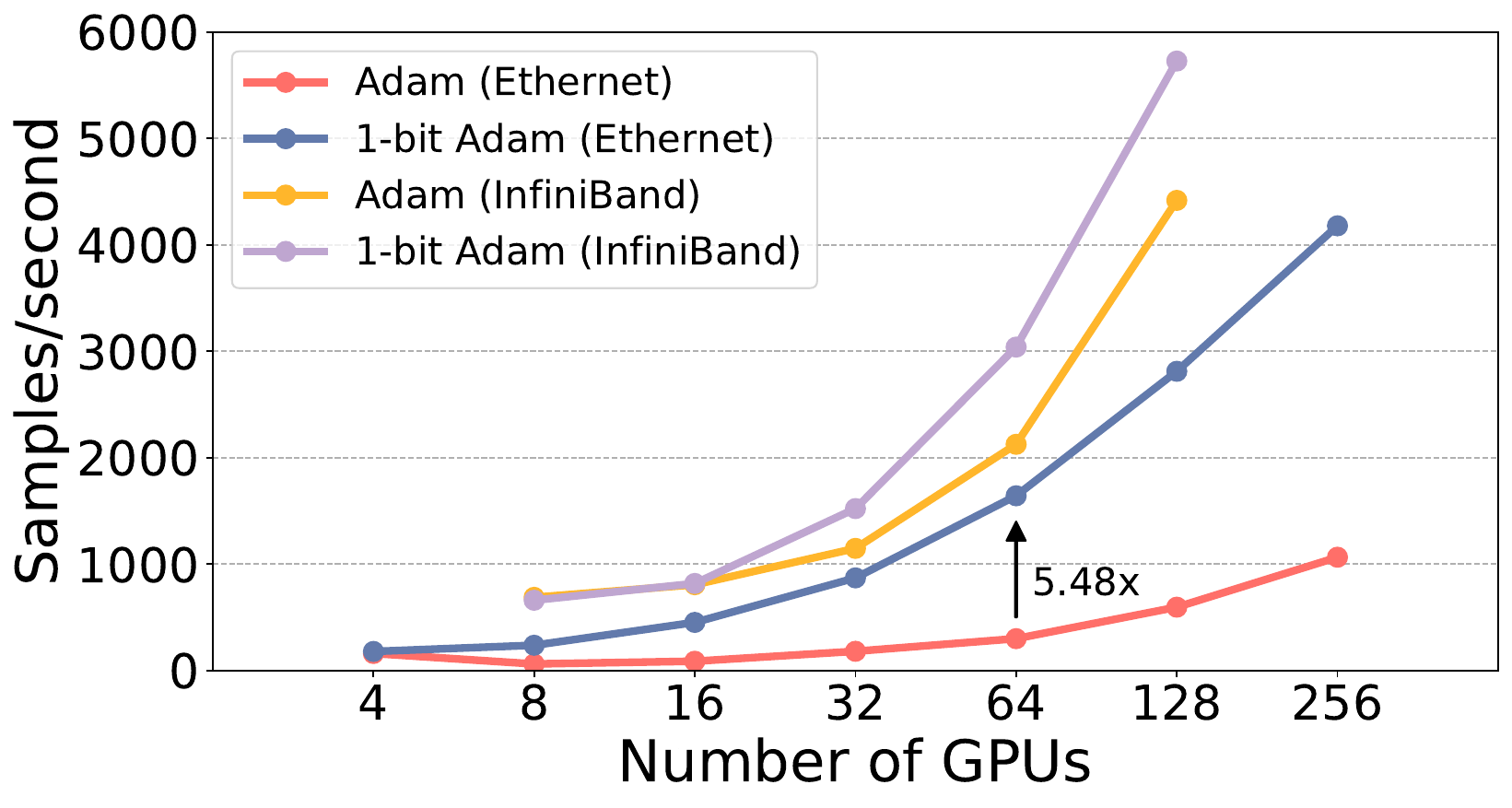}\label{fig:e2e-1}
\end{minipage}
}
\subfigure[Bert-Large pre-training, batch size = 4K]{
\begin{minipage}[t]{0.34\linewidth}
\centering
\includegraphics[width=1\textwidth]{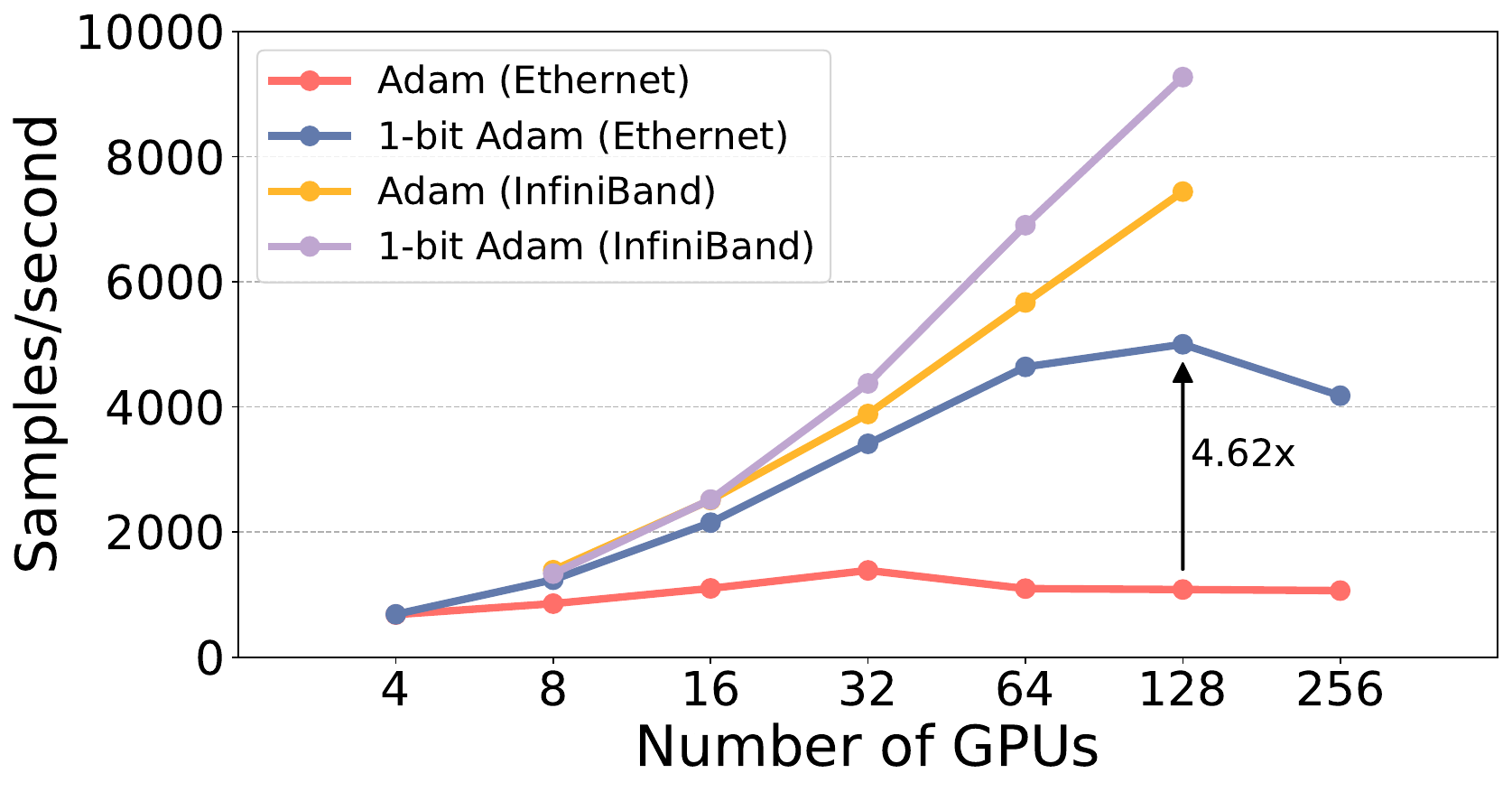}\label{fig:e2e-2}
\end{minipage}
}
\subfigure[SQuAD fine-tuning, batch size = number of GPUs $\times$ 3]{
\begin{minipage}[t]{0.28\linewidth}
\centering
\includegraphics[width=1\textwidth]{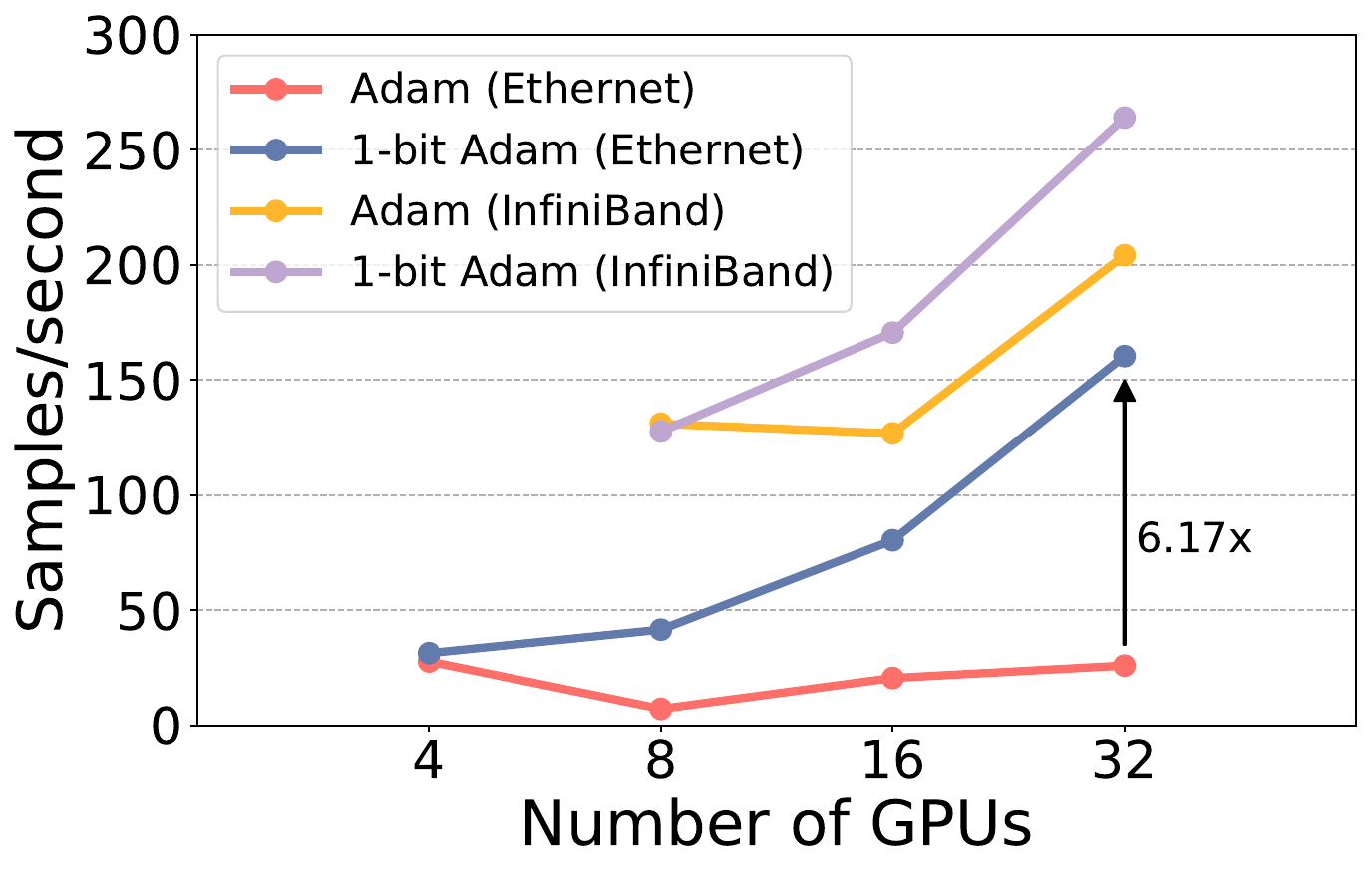}\label{fig:e2e-3}
\end{minipage}%
}
\centering
\caption{Scalability of {\OA} for BERT-Large pre-training sequence length 128 and SQuAD 1.1 fine-tuning on V100 GPUs. {Adam} lines represent the throughput at {\OA}'s warmup stage (i.e., baseline {Adam}'s throughput). {\OA} lines represent the throughput at compression stage. Annotations represent the highest speedup achieved in each figure. Note that this is the speedup between warmup and compression stage. The end-to-end speedup also depends on the percentage of warmup.}\label{fig:e2e}\vspace{-0.5cm}
\end{figure*}

% \begin{figure}[t]
% \centering
% \includegraphics[width=0.45\textwidth]{new_bandwidth.pdf}
% \caption{Per-iteration speedup (on $64$ V100 GPUs) for BERT-Base under different network bandwidths.}\label{fig:bert_speed}\vspace{-0.4cm}
% \end{figure}

\paragraph{Performance results}
Computed as 1/(warmup ratio + (1 - warmup ratio)/16) for FP16 training, {\OA} offers up to 5x less end-to-end communication volume for BERT-Base and BERT-Large. This leads to to 3.3x higher throughput for BERT-Large sequence length 128 pre-training and up to 2.9x higher throughput for SQuAD fine-tuning. This end-to-end throughput improvement is enabled by the 5.48x (Figure~\ref{fig:e2e-1}) and 6.17x (Figure~\ref{fig:e2e-3}) speedup observed during the compression stage. Figure~\ref{fig:e2e-2} shows that {\OA} also provides better scalability: {Adam}'s throughput reaches peak at 32 GPUs on Ethernet, while {\OA}'s throughput keeps increasing until 128 GPUs. It is also worth mentioning that {\OA} on Ethernet (4.1 Gbps effective bandwidth, 4 GPUs per node) is able to achieve comparable throughput as {Adam} on InfiniBand (near 100 Gbps effective bandwidth, 8 GPUs per node), which demonstrates {\OA}'s efficiency considering the hardware differences.

In Figure~\ref{fig:bert-time} we also measured the total training time of BERT-Large pre-training seqlen 128 when using batch size $4K$ on 64 GPUs on the Ethernet cluster. It takes 174.3 hours for baseline Adam to complete the training, while {\OA} only needs 51.5 hours. This 3.4x speedup is consistent with the speedup computed based on the throughput analysis above.

% In Figure~\ref{fig:bert_speed}, we compress the communication data from 32bits to 1bit for {\OA} and report the per-iteration speedup under different network conditions. Specifically, we use traffic control utility \textit{tc} to shape the bandwidth from 100Gbits to 100Mbits on InfiniBand. With the network going slow, the compressed case can achieve a stable speedup by up to 22$\times$ over the uncompressed case ($10\times$ speedup at 2Gbits bandwidth and $3\times$ at 10Gbits bandwidth).

\subsection{ResNet on CIFAR10 and ImageNet}\label{resnet}

\begin{figure}[t]
\centering
\subfigure[Training loss]{
\begin{minipage}[t]{0.53\linewidth}
\centering
\includegraphics[width=1\textwidth]{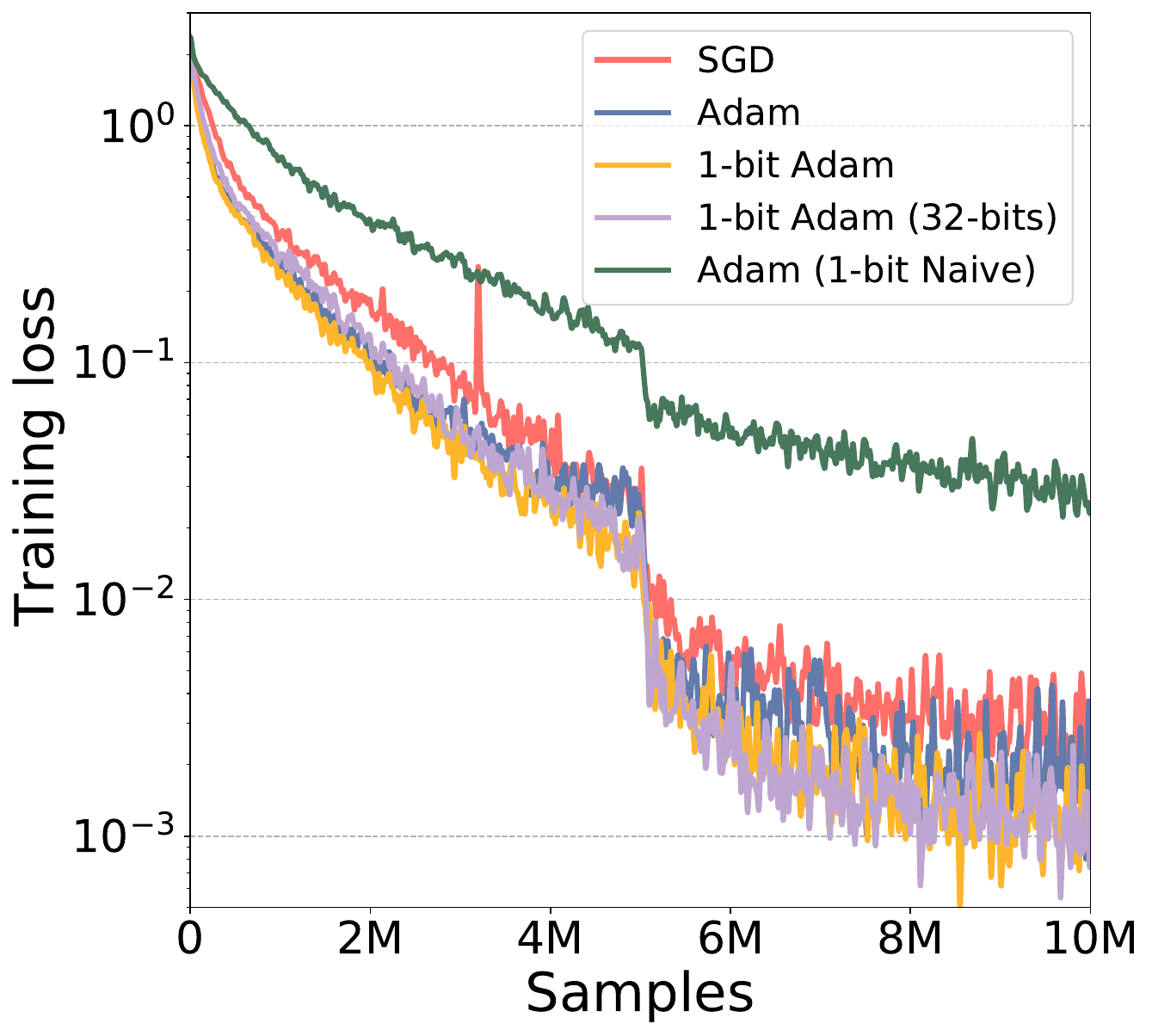}
\end{minipage}
}
\subfigure[Testing accuracy]{
\begin{minipage}[t]{0.37\linewidth}
\centering
\includegraphics[width=1\textwidth]{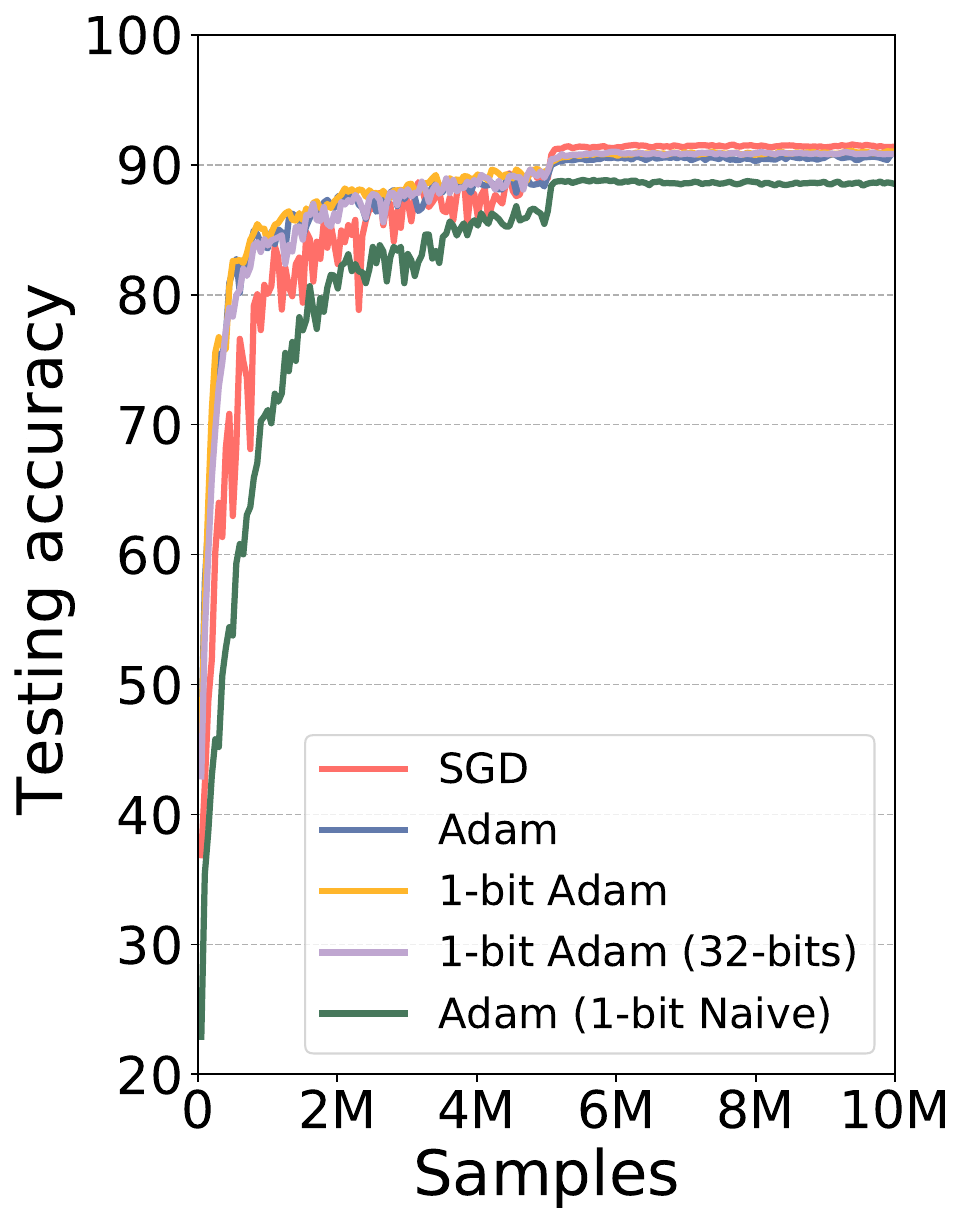}
\end{minipage}
}
\centering
\caption{Sample-wise convergence speed for ResNet-18.}\label{fig:resnet}
\end{figure}

To further evaluate the convergence speed of {\OA} and related works, we train CIFAR10 using ResNet-18\citep{7780459}. The dataset has a training set of 50000 images and a test set of 10000 images, where each image is given one of the 10 labels. We run the experiments on $8$ 1080Ti GPUs where each GPU is used as one worker. The batch size on each worker is $128$ and the total batch size is $1024$.

We evaluate five implementations for comparison: 1) Original {SGD}. 2) Original {Adam} \citep{adam}. 3) {\OA} where we use $13$ out of $200$ epochs as warmup. 4)  \textbf{{\OA} (32-bits)} where we do not compress the momentum while still freezing the variance. 5) \textbf{Adam(1-bit Naive)} where we compress the gradient instead of momentum, and don't freeze the variance. We set the learning rate as $1\times 10^{-1} $ for {SGD} and $1\times 10^{-4}$ for the other 4 cases. For all five cases, the learning rate is decayed into $10\%$ of the original after every $100$ epochs.

As illustrated in Figure~\ref{fig:resnet}, {\OA} achieves similar convergence speed as {Adam} and {\OA} {(32-bits)}. {SGD} has a slightly slower convergence speed while {Adam(1-bit Naive)} is much worse. This and Section~\ref{sec:moti-convergence} demonstrate that existing compression method doesn't work for {Adam}. In the supplementary materials we further compare {\OA} with other related works using ResNet-18.

\begin{figure}[t]
\centering
\includegraphics[width=0.45\textwidth]{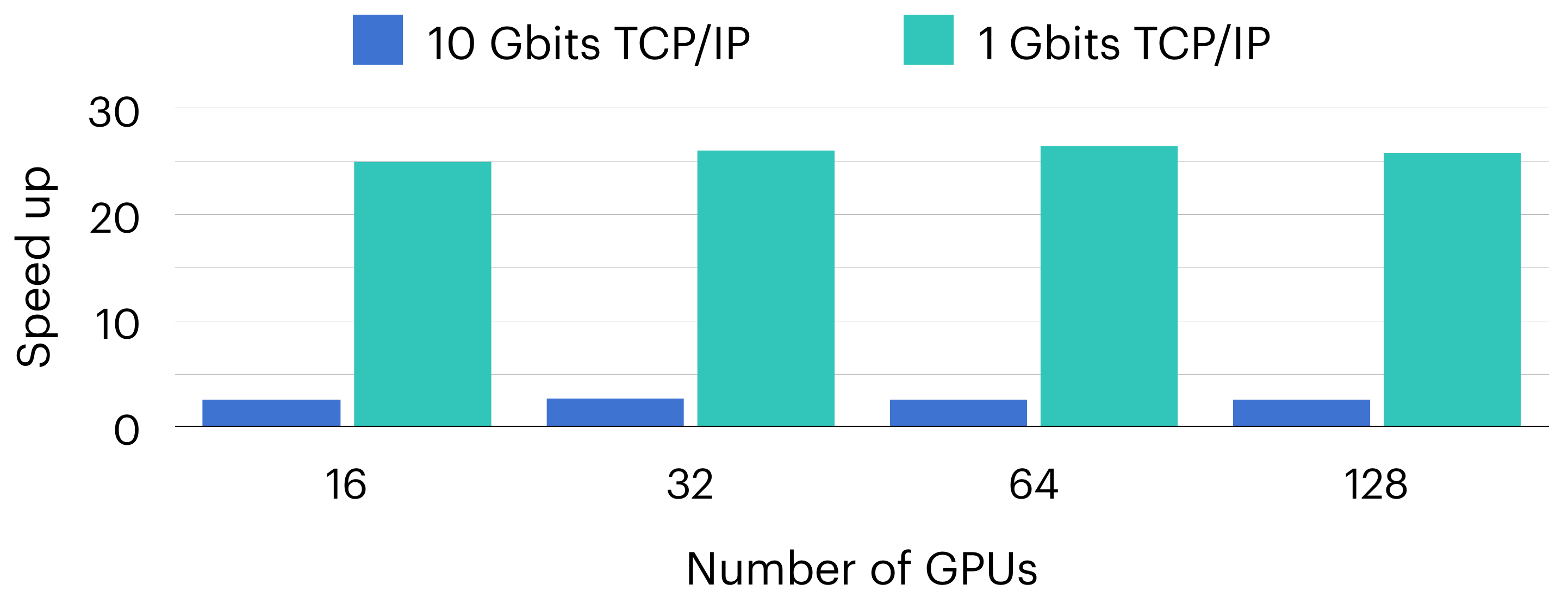}
\caption{Speedup of ResNet-152 on ImageNet. Each server has 8 V100 GPUs interconnected by NVLink, servers are connected by 10Gbits or 1Gbits TCP/IP network.}
\label{fig:resnet_speedup}
\end{figure}

Moreover, to see how {\OA} could speedup the training in this case, we report speedup  results of training ResNet-152 on ImageNet~\citep{imagenet} using different numbers of GPUs, in Figure~\ref{fig:resnet_speedup}. As we can see that {\OA} could potentially speedup the training especially when the bandwidth is limited.

\subsection{Deep Convolutional Generative Adversarial Networks}

\begin{figure}[t]
\centering
\subfigure[Training loss]{
\begin{minipage}[t]{0.33\linewidth}
\centering
\includegraphics[width=1\textwidth]{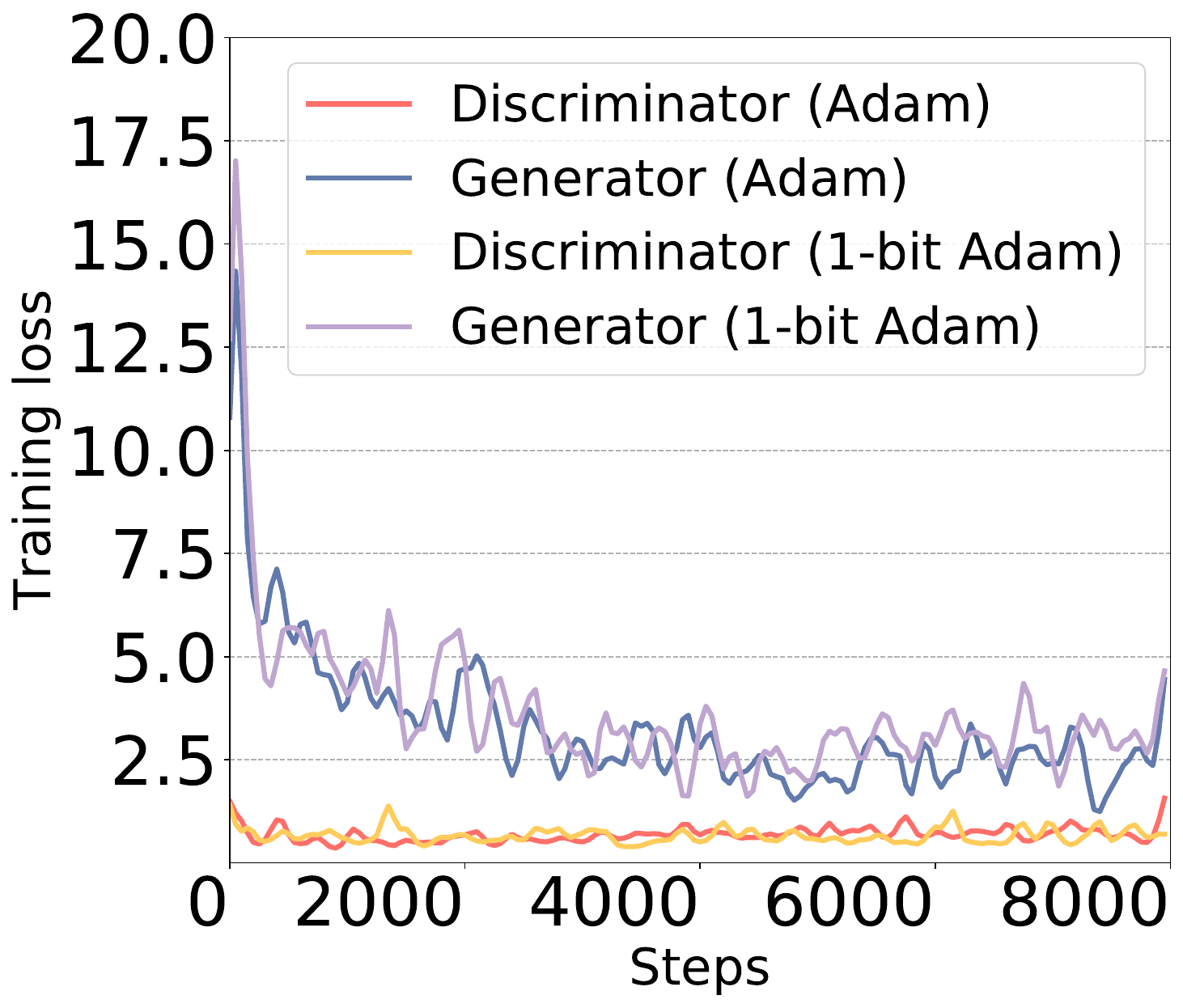}
\end{minipage}
}
\subfigure[Generated images (Adam)]{
\begin{minipage}[t]{0.27\linewidth}
\centering
\includegraphics[width=1\textwidth]{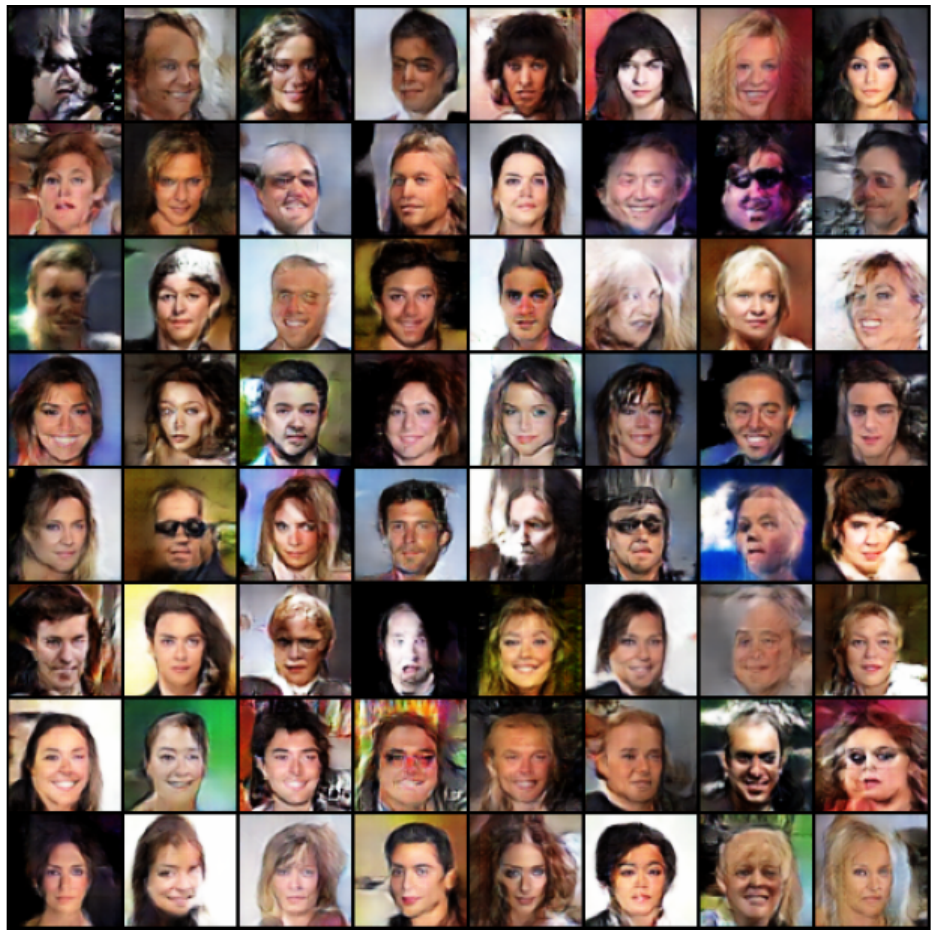}
\end{minipage}
}
\subfigure[Generated images (1-bit Adam)]{
\begin{minipage}[t]{0.27\linewidth}
\centering
\includegraphics[width=1\textwidth]{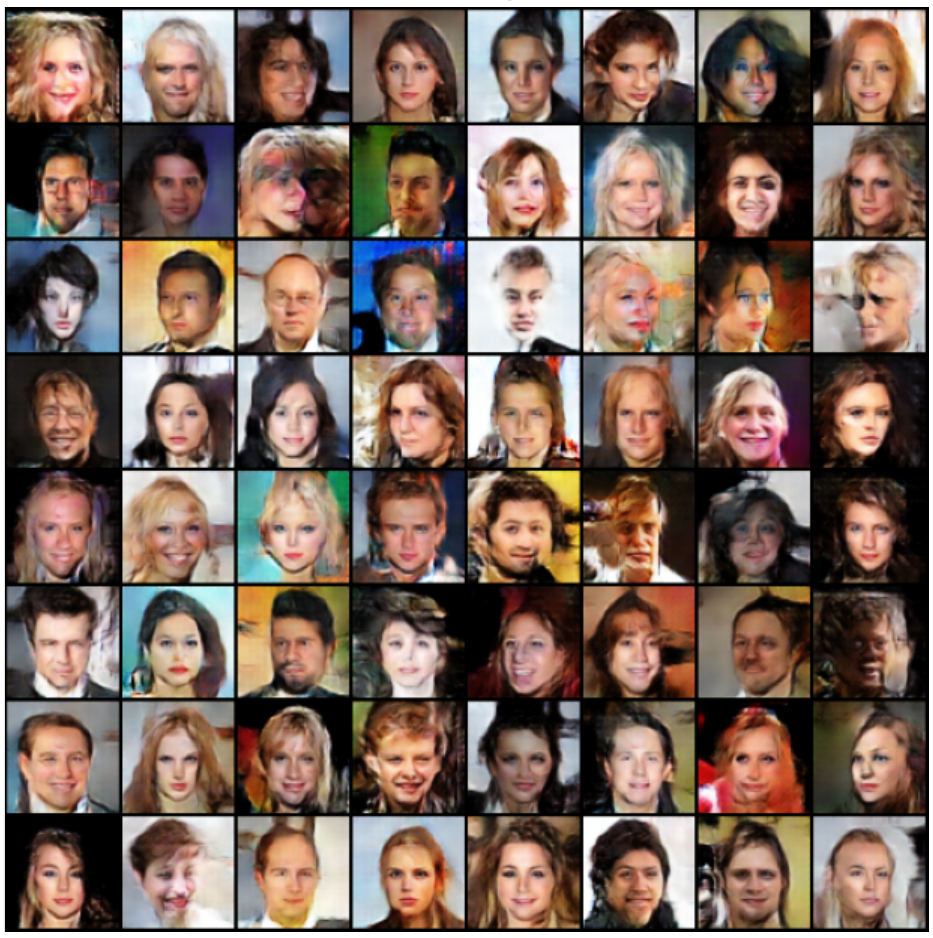}
\end{minipage}
}
\centering
% \vspace{-0.4cm}
% \caption{Performance of Adam and 1-bit Adam for training GAN. The training loss result is included in (a); (b) are the fake images generated using Adam, and (c) are the fake images generated using 1-bit Adam ($20\%$ of steps are used as warm-up)}\label{fig:gan}
\caption{Comparison of Adam and 1-bit Adam ($20\%$ warmup steps) for training Deep Convolutional Generative Adversarial Networks (DCGAN).}
\label{fig:gan}
% \vspace{-0.3cm}
\end{figure}

To further understand the correctness of {\OA} on more tasks, we apply it to the training of Generative Adversarial Networks (GAN). We choose Deep Convolutional GAN \cite{radford2015unsupervised} as the model, which adopts convolutional and convolutional-transpose layers for the discriminator and generator. We use CelebFaces Attributes Dataset (CelebA) \cite{liu2015faceattributes} as the training data, which contains more than 200K celebrity images. The task is to train the discriminator and generator in an adversarial way, such that the generator can create fake but vivid face images. Figure \ref{fig:gan} shows the training loss and generated images by using original Adam optimizer and {\OA}. The results show that {\OA} can achieve almost the same training accuracy as the Adam optimizer.

\section{Conclusions}
In this paper, we propose an error-compensated {Adam} preconditioned momentum SGD algorithm, {\OA}, which provides both communication efficiency and {Adam}'s convergence speed. Our theoretical analysis demonstrates that {\OA} admits a linear speed w.r.t the number of workers in the network, and is robust to any compression method. We validate the performance of {\OA} empirically on BERT pre-training/fine-tuning, ResNet training and DCGAN training tasks on up to 256 GPUs. Results show that {\OA} provides same sample-wise convergence speed as uncompressed {Adam}, reduces communication volume by up to 5x, and runs up to 3.3 times faster than uncompressed algorithms. Beyond those results, it's interesting to see the performance of {\OA} on wider variety of tasks, e.g., reinforcement learning, which we leave for the future work.

\newpage
\bibliographystyle{abbrvnat}
\bibliography{reference}

\onecolumn
{\Huge \centering Supplementary}
\section{Bert Pre-training: speedup under different network bandwidths}
\begin{figure}[h]
\centering
\includegraphics[width=0.5\textwidth]{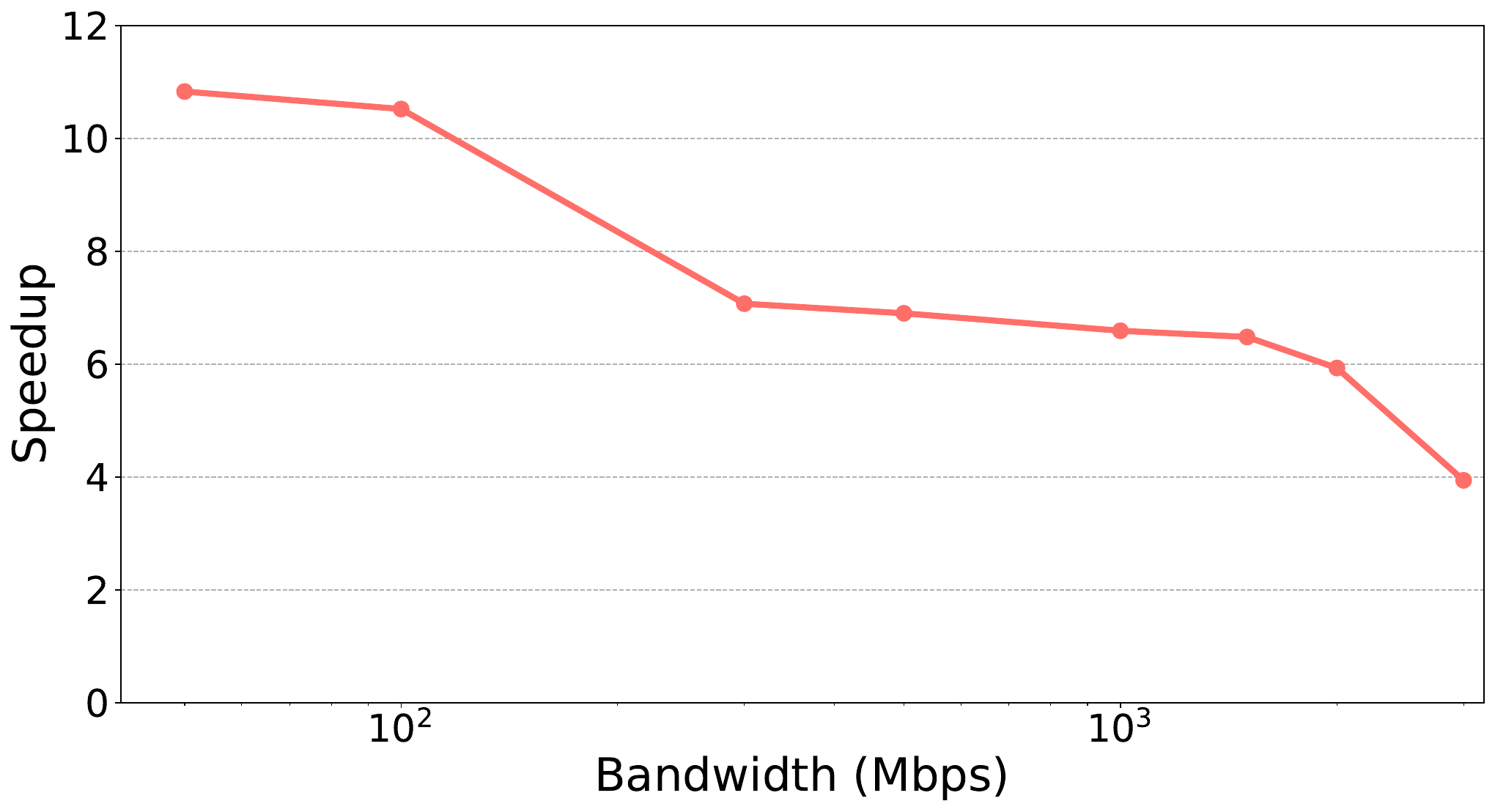}
\caption{Throughput speedup (on $256$ V100 GPUs) between Adam and {\OA} compression stage for BERT-Large pre-training under different network bandwidths, from 50Mbits to 3Gbits. The x-axis is in log scale.}\label{fig:bert_speed}\vspace{-0.1cm}
\end{figure}

In Figure~\ref{fig:bert_speed}, we evaluate the throughput speedup between Adam and {\OA} compression stage under different network conditions. Specifically, we use traffic control utility \textit{tc} to shape the bandwidth from 50Mbits to 3Gbits on Ethernet. With the network going slow, {\OA} compression stage can achieve a speedup up to $10.83\times$ over the uncompressed Adam ($6.59\times$ speedup at 1Gbits bandwidth and $5.93\times$ at 2Gbits bandwidth).

\section{ResNet: comparing with additional communication efficient algorithms}
Beyond Adam, there have been many communication efficient optimization algorithms proposed for SGD and Momentum SGD. We also compare the convergence speed of those algorithms with {\OA} for training ResNet-18 on CIFAR10.

We evaluate five implementations for comparison:
\begin{enumerate}
\item \textbf{DoubleSqueeze}~\citep{tangdouble}: In DoubleSqueeze, the stochastic gradient $\g_t$ is compressed with error compensation. We use $1$-bit compression here.
\item \textbf{Momentum SGD}: The updating rule of Momentum SGD admits 
\begin{align*}
\m_{t+1} =& \beta \m_t + (1-\beta)\g_t,\\
\x_{t+1} = & \x_t - \gamma\m_{t+1},
\end{align*}
where $\g_t$ is the stochastic gradient and $\m_t$ is the momentum. 
\item \textbf{Error-Feedback Momentum SGD (EF Momentum SGD)}~\citep{NIPS2019_9321}: This algorithm is similar with \citet{NIPS2019_9321}, where the momentum $\m_t$ is being compressed with error compensation. We use $1$-bit compression here.
\item \textbf{Local SGD}~\citep{stich2019local}: In local SGD, the model would get updated using local gradient following SGD, and after every $\tau$ step, the model would be averaged over workers.
\item \textbf{Local SGD with Momentum}: In local SGD, the model would get updated using local gradient following Momentum SGD, and after every $\tau$ step, both model and momentum would be averaged over workers.
\end{enumerate}

We set the momentum $\beta=0.9$ here. In order to get the comparable communication reduction, we set $\tau=4$ for Local SGD. The learning rate is grid searched from $\gamma\in\{0.5,0.1,0.01,0.001\}$ for each algorithm except Adam and {\OA} (in which we set learning rate $\gamma=1\times 10^{-4}$), and we find that $\gamma=0.1$ works best for all of them.

\begin{figure}[t]
\centering
\subfigure[Training loss]{
\begin{minipage}[t]{0.56\linewidth}
\centering
\includegraphics[width=1\textwidth]{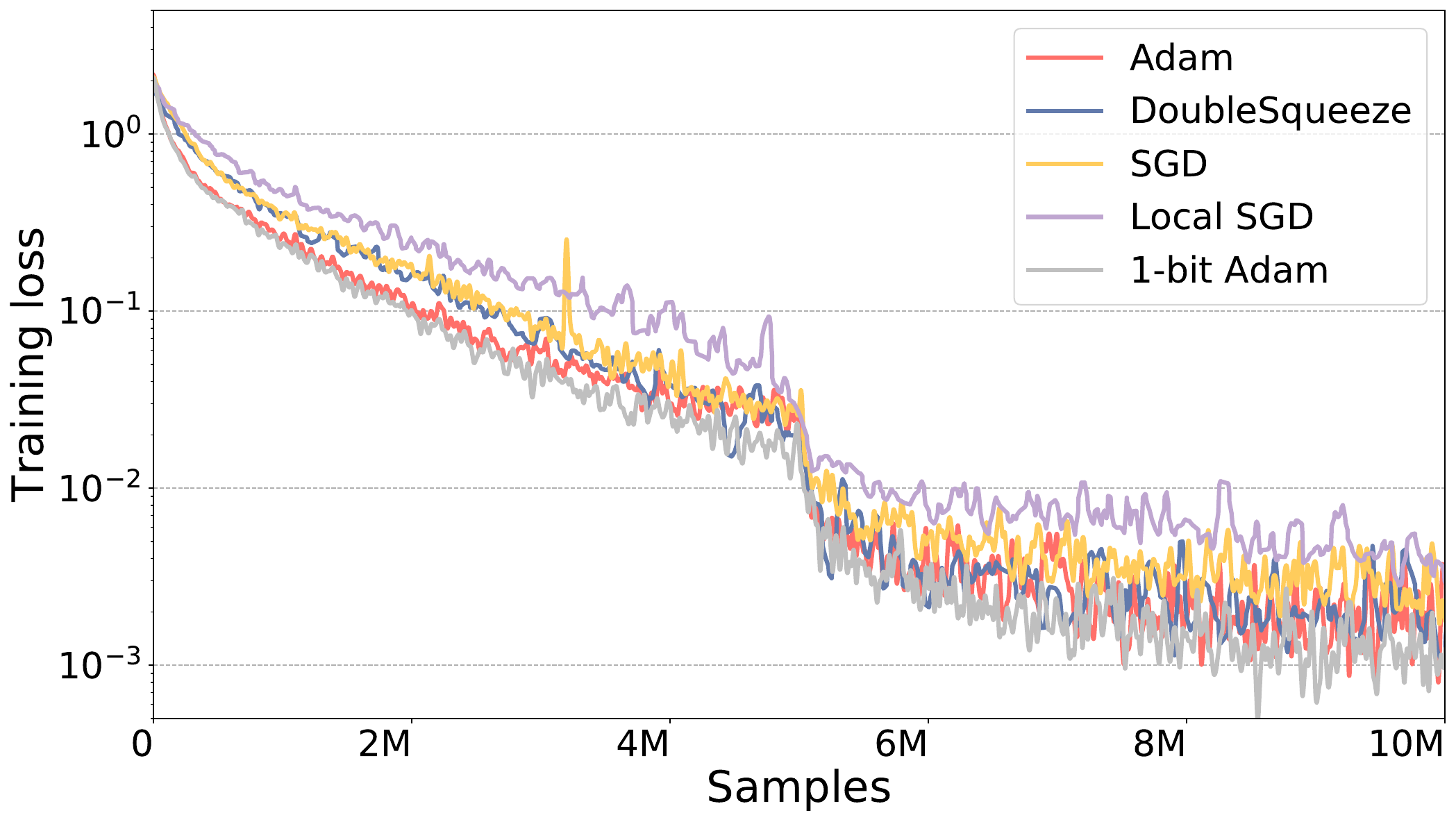}
\end{minipage}
}
\subfigure[Testing accuracy]{
\begin{minipage}[t]{0.38\linewidth}
\centering
\includegraphics[width=1\textwidth]{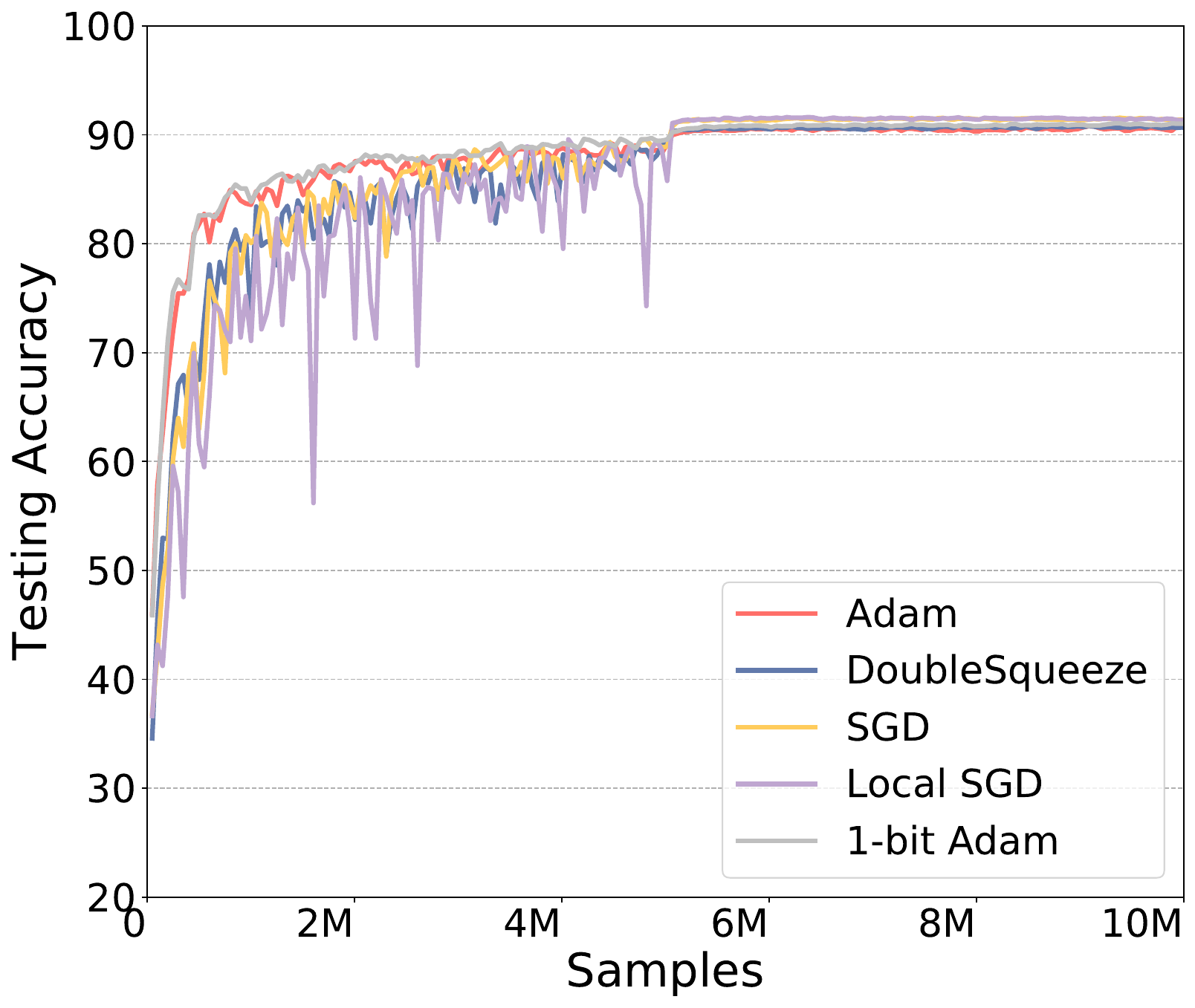}
\end{minipage}
}
\centering
\caption{Epoch-wise convergence speed for ResNet-18. We compare {\OA} with SGD-type of communication efficient algorithms. }\label{fig:supp_mom0_resnet}
\end{figure}

\begin{figure}[t]
\centering
\subfigure[Training loss]{
\begin{minipage}[t]{0.56\linewidth}
\centering
\includegraphics[width=1\textwidth]{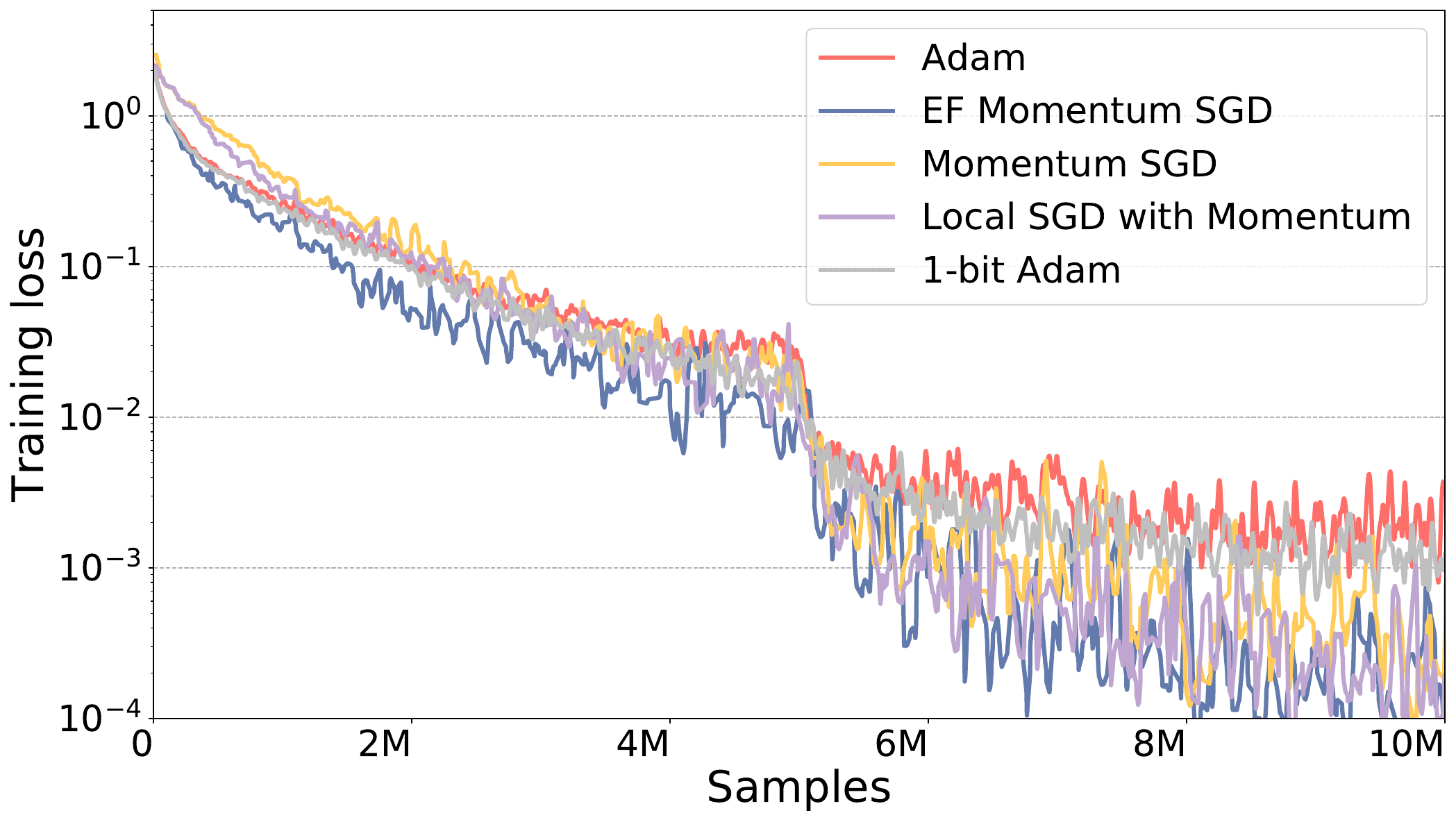}
\end{minipage}
}
\subfigure[Testing accuracy]{
\begin{minipage}[t]{0.38\linewidth}
\centering
\includegraphics[width=1\textwidth]{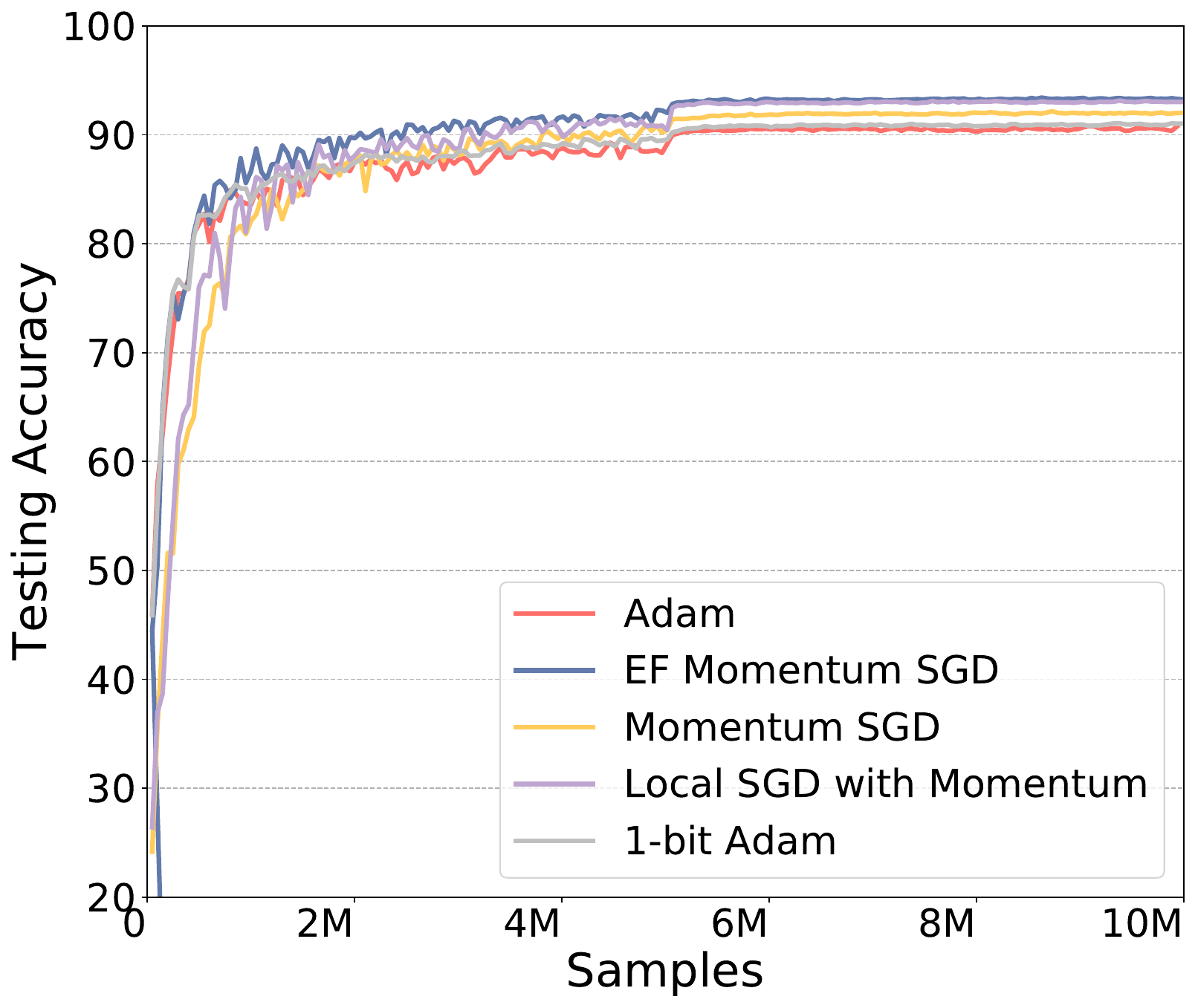}
\end{minipage}
}
\centering
\caption{Epoch-wise convergence speed for ResNet-18. We compare {\OA} with Momentum SGD-type of communication efficient algorithms, and the momentum is set to be $\beta=0.9$.}\label{fig:supp_mom90_resnet}
\end{figure}

In Figure~\ref{fig:supp_mom0_resnet} we compare the convergence speed of {\OA} with DoubleSqueeze and Local SGD, and in Figure~\ref{fig:supp_mom90_resnet}, we compare the convergence speed of {\OA} with EF Momentum SGD and Local SGD with Momentum.

Notice that for training ResNet-18, both EF Momentum SGD and Local Momentum SGD admits a faster convergence speed than {\OA}, this is because the uncompressed Momentum SGD runs faster than Adam.

\begin{figure}[t]
\centering
\subfigure[Training loss]{
\begin{minipage}[t]{0.45\linewidth}
\centering
\includegraphics[width=1\textwidth]{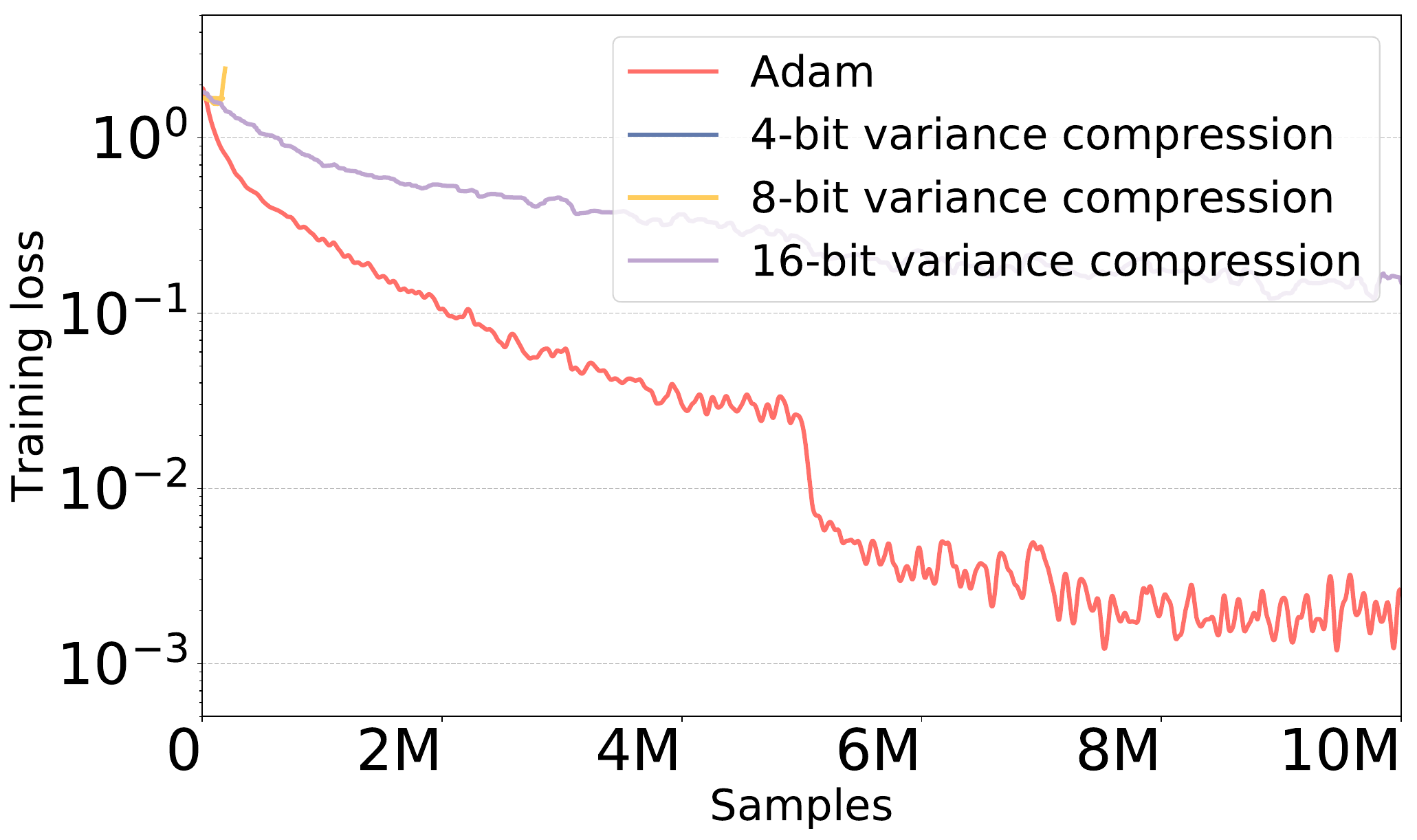}
\end{minipage}
}
\subfigure[Testing accuracy]{
\begin{minipage}[t]{0.45\linewidth}
\centering
\includegraphics[width=1\textwidth]{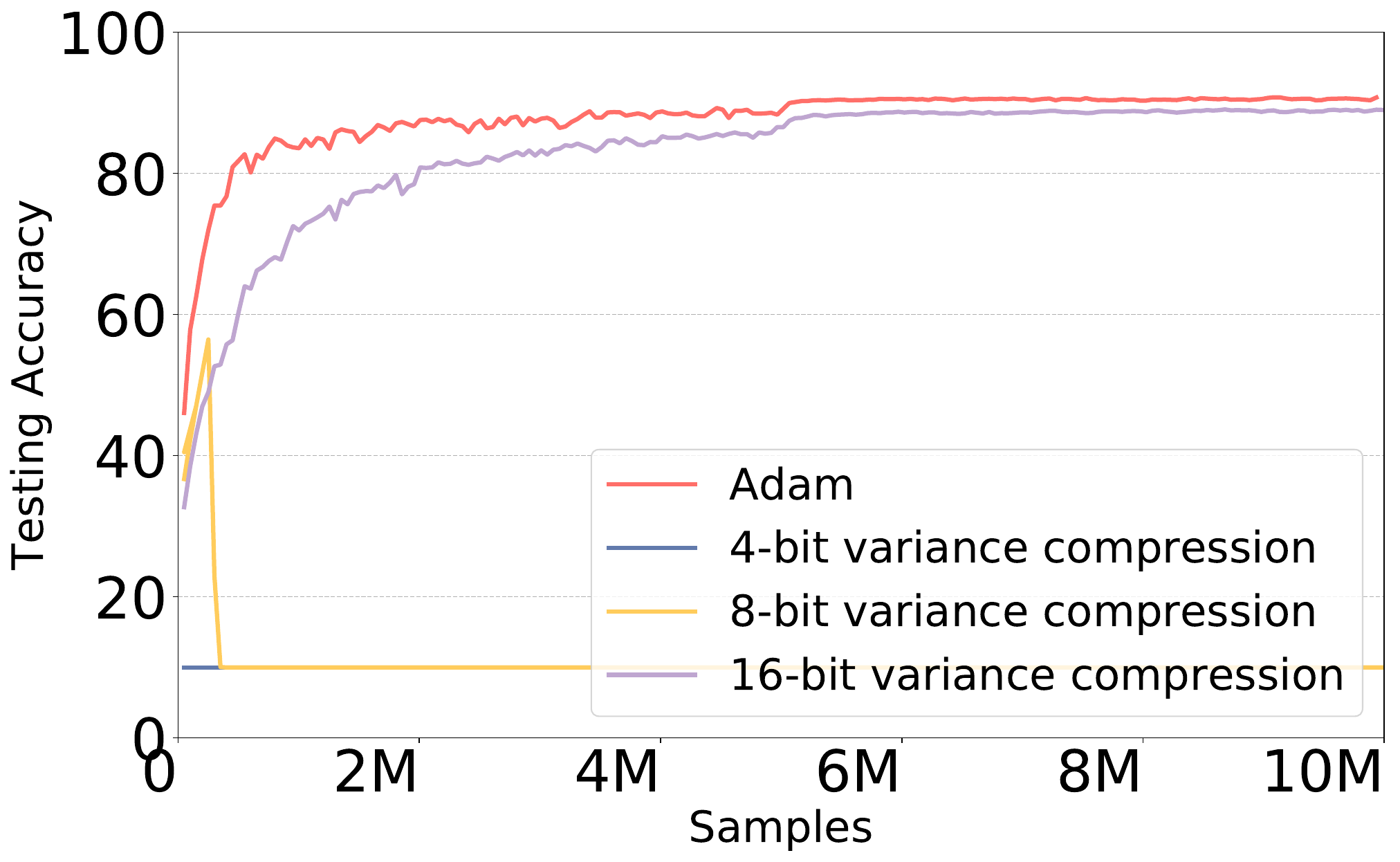}
\end{minipage}
}
\vspace{-0.4cm}
\centering
\caption{ResNet-18 on CIFAR10. Momentum and variance are compressed into 1-bit and $n$-bit. When $n\leq 8$, the training cannot converge, so we do not include the result above.
}
\label{fig:nbits_variance}
\end{figure}

\begin{figure}[t]
\centering
\subfigure[Training loss]{
\begin{minipage}[t]{0.45\linewidth}
\centering
\includegraphics[width=1\textwidth]{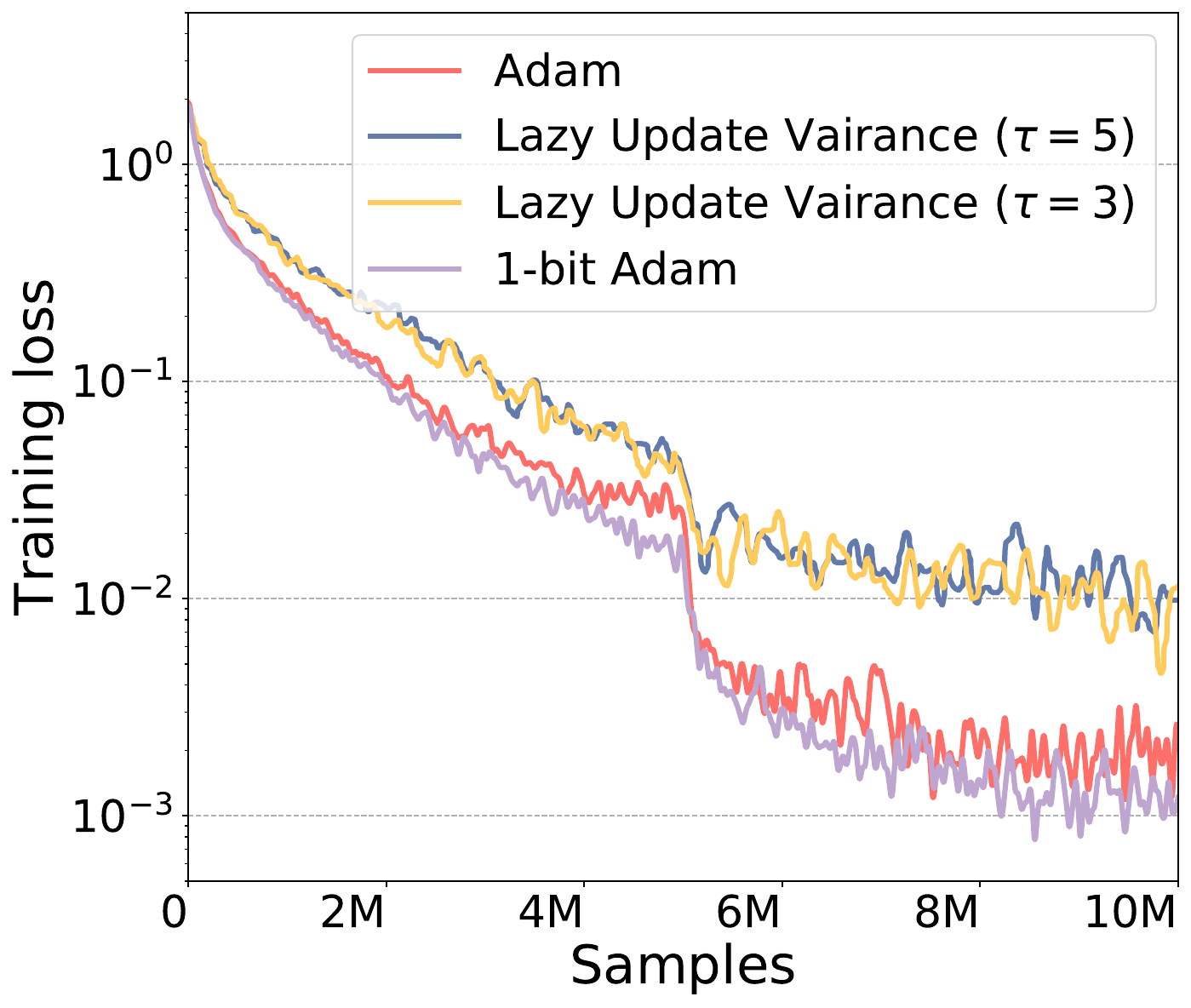}
\end{minipage}
}
\subfigure[Testing accuracy]{
\begin{minipage}[t]{0.45\linewidth}
\centering
\includegraphics[width=1\textwidth]{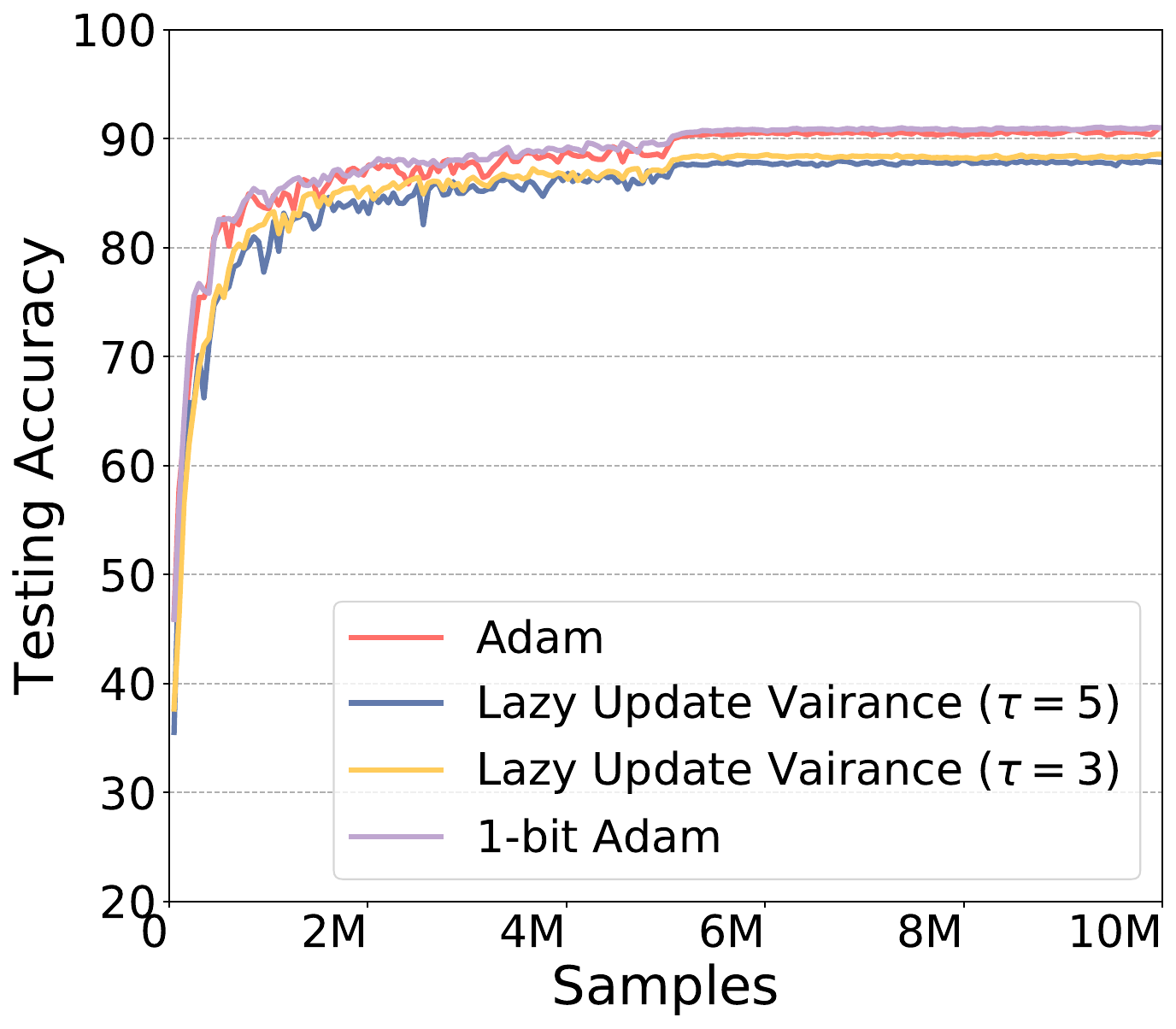}
\end{minipage}
}
\vspace{-0.4cm}
\centering
\caption{ResNet-18 on CIFAR10. The variance term would get averaged after every $\tau$ stpes.
}
\label{fig:lazy_variance}
\end{figure}
Meanwhile, for Adam, we also evaluate the influence of the variance term to the convergence speed with some following tryouts:
\begin{itemize}
\item \textbf{Adam with $n$-bits Variance Compression}: This algorithm would allreduce both the momentum term and variance term, with variance term being compressed into $n$-bits representation \citep{Alistarh2017-yh}. This design is to see whether we could still achieve comparable convergence speed with the variance term being compressed. The results is in Figure~\ref{fig:nbits_variance}.
\item \textbf{Adam with Lazily Updated Variance}: Here we only allreduce the variance term without compression after every $\tau$ steps, and the the variance term would get updated continuously using local gradients. The results is in Figure~\ref{fig:lazy_variance}.
\end{itemize}
Unfortunately, both methods fail to achieve a comparable convergence speed with Adam. Therefore the 1-bit Adam method proposed in the paper is the only solution we found that could provide the same convergence speed with vanilla Adam.

\section{Proof to the updating form}
Since our algorithm is equivalent to running a parameter-server prototype communication on each chunk of the gradient, so below we will assume a parameter-server model (which means the tensor is not required to be divided into $n$ chunks) for simplicity.

According to the algorithm description in Section~\ref{alg:description}, at iteration $t+1$, the updating step of the momentum term $\bm{m}_{t+1}$ can be divided into two steps:
\begin{enumerate}
\item Local Update and Compress: each worker locally update $\bm{m}_t$ and use the error-compensate strategy for compressing.
\begin{align*}
\m_{t}^{(i)} =& \beta \bm{m}_t +  (1-\beta)\bm{g}_t^{(i)}\\
\m_{t+\frac{1}{2}}^{(i)} = & \C_{\omega}[\m_{t}^{(i)} + \bm{\delta}_t^{(i)}]\\
\bm{\delta}_{t+1}^{(i)} = & \m_{t}^{(i)} + \bm{\delta}_t^{(i)} - \m_{t+\frac{1}{2}}^{(i)}.
\end{align*}
\item All workers send its  $\m_{t+\frac{1}{2}}^{(i)}$ to the server. The server takes the average over them and compress it again using error-compensation.
\begin{align*}
\m_{t+\frac{1}{2}} =& \frac{1}{n}\sum_{i=1}^n \m_{t+\frac{1}{2}}^{(i)}\\
\m_{t+1} = & \C_{\omega}[\m_{t+\frac{1}{2}} + \bm{\delta}_t]\\
\bm{\delta}_{t+1} = & \m_{t+\frac{1}{2}} + \bm{\delta}_t - \m_{t+1}.
\end{align*}
\item The server broadcast $\m_{t+1}$ to all workers, and all workers update the local model according to
\begin{align*}
\bm{x}_{t+1} = \bm{x}_{t} - \bm{\gamma}\m_{t+1}\oslash \sqrt{\bm{v}_{T_{w}}^2}.
\end{align*}

\end{enumerate}
So actually the updating rule above can be summarized as
\begin{align*}
\m_{t+1} =& \C_{\omega}[\m_{t+\frac{1}{2}} + \bm{\delta}_t]\\
= & \m_{t+\frac{1}{2}} + \bm{\delta}_t - \bm{\delta}_{t+1}\quad\text{(from the definition of $\bm{\delta}_{t+1}$)}\\
= & \frac{1}{n}\sum_{i=1}^n \C_{\omega}[\m_{t}^{(i)} + \bm{\delta}_t^{(i)}]+ \bm{\delta}_t - \bm{\delta}_{t+1}\\
= & \frac{1}{n}\sum_{i=1}^n\left( \m_{t}^{(i)} + \bm{\delta}_t^{(i)} - \bm{\delta}_{t+1}^{(i)} \right) + \bm{\delta}_t - \bm{\delta}_{t+1}\quad\text{(from the definition of $\bm{\delta}_{t+1}^{(i)}$)}\\
= & \beta \m_t + \frac{1-\beta}{n}\sum_{i=1}^n \bm{g}_t^{(i)} + \left( \frac{1}{n}\sum_{i=1}^n \bm{\delta}_t^{(i)} + \bm{\delta}_t\right) - \left( \frac{1}{n}\sum_{i=1}^n \bm{\delta}_{t+1}^{(i)} + \bm{\delta}_{t+1}\right).
\end{align*}
Denote
\begin{align*}
\overline{\bm{g}}_t = & \frac{1}{n}\sum_{i=1}^n \bm{g}_t^{(i)}\\
\overline{\bm{\delta}}_{t} = & \frac{1}{n}\sum_{i=1}^n \bm{\delta}_t^{(i)} + \bm{\delta}_t,
\end{align*}
the update rule of $\m_{t}$ can be summarized as
\begin{align*}
\m_t = \beta\m_{t-1} + (1-\beta)\overline{\bm{g}}_t + \overline{\bm{\delta}}_{t-1} - \overline{\bm{\delta}}_{t},
\end{align*}
and
\begin{align*}
\x_{t+1} = \x_t - \gamma V\m_t,
\end{align*}
where $V= \text{diag}(1/\sqrt{v_1},1/\sqrt{v_2},\cdots,1/\sqrt{v_d})$ is the a diagonal matrix that spanned with $\v_{T_{w}}$.

\section{Proof to Theorem~\ref{theo:global}}
Notice that in for {\OA}, the learning rate for each coordinate is different. In order to simplify our analysis, we instead consider another function that is defined as
\begin{align*}
H(\z) = F(V^{\frac{1}{2}}\z),
\end{align*}
also
\begin{align*}
h(\z) = f(V^{\frac{1}{2}}\z),
\end{align*}
where ${V} $ is a diagonal matrix.

In this case we have
\begin{align*}
{V}^{\frac{1}{2}}\nabla f({V}^{\frac{1}{2}}\z) = \nabla h(\z).
\end{align*}
Therefore the updating rule of {\OA} in the view of $h(\cdot)$ is
\begin{align*}
{V}^{\frac{1}{2}}\z_{t+1} = {V}^{\frac{1}{2}}\z_t - \gamma {V}^{\frac{1}{2}}\left({V}^{\frac{1}{2}}\m_t\right).
\end{align*}
It can be easily verified that
\begin{align*}
\bm{m}_t =& (1-\beta) \sum_{s=0}^t \beta^{t-s} \overline{\bm{g}}_s + \sum_{s=0}^t \beta^{t-s}( \overline{\bm{\delta}}_{s-1} - \overline{\bm{\delta}}_{s})\\
= & (1-\beta) \sum_{s=0}^t \beta^{t-s} \frac{1}{n}\sum_{i=1}^n\nabla F(V^{\frac{1}{2}}\z_t;\xi_t^{(i)}) + \sum_{s=0}^t \beta^{t-s}( \overline{\bm{\delta}}_{s-1} - \overline{\bm{\delta}}_{s})
\end{align*}
which means
\begin{align*}
{V}^{\frac{1}{2}}\bm{m}_t =& (1-\beta) \sum_{s=0}^t \beta^{t-s} \frac{1}{n}\sum_{i=1}^n{V}^{\frac{1}{2}}\nabla F(V^{\frac{1}{2}}\z_t;\xi_t^{(i)}) + \sum_{s=0}^t \beta^{t-s}{V}^{\frac{1}{2}}( \overline{\bm{\delta}}_{s-1} - \overline{\bm{\delta}}_{s})\\
= & (1-\beta) \sum_{s=0}^t \beta^{t-s} \frac{1}{n}\sum_{i=1}^n\nabla H(V^{\frac{1}{2}}\z_t;\xi_t^{(i)}) + \sum_{s=0}^t \beta^{t-s}{V}^{\frac{1}{2}}( \overline{\bm{\delta}}_{s-1} - \overline{\bm{\delta}}_{s})\\
= & (1-\beta) \sum_{s=0}^t \beta^{t-s} \overline{\bm{g}}_s(\z) + \sum_{s=0}^t \beta^{t-s}{V}^{\frac{1}{2}}( \overline{\bm{\delta}}_{s-1} - \overline{\bm{\delta}}_{s}),
\end{align*}
where $\overline{\bm{g}}_s(\z)$ is the corresponding averaged stochastic gradient computed in the view of loss function $h(\cdot)$.

Then, if we define $\m_t(\z) = {V}^{\frac{1}{2}}\m_t$, the updating rule of $\m_t(z)$ admits
\begin{align*}
\m_t(\z) = \beta\m_{t-1}(\z) + (1-\beta)\overline{\bm{g}}_t(\z) + {V}^{\frac{1}{2}}\overline{\bm{\delta}}_{t-1} - {V}^{\frac{1}{2}}\overline{\bm{\delta}}_{t}, \numberthis\label{supp:trans_eq1}
\end{align*}
and
\begin{align*}
{V}^{\frac{1}{2}}\z_{t+1} =& {V}^{\frac{1}{2}}\z_t - \gamma {V}^{\frac{1}{2}}\m_t(\z)\\
\z_{t+1} =& \z_t - \gamma \m_t(\z).\numberthis\label{supp:trans_eq2}
\end{align*}
From \eqref{supp:trans_eq1} and \eqref{supp:trans_eq2} we shall see that using different learning rate for each coordinate is equivalent to optimizing a new loss function defined on  scaling the original coordinate and using a uniform learning for all coordinates. Therefore below we first study the behavior of the error-compensated momentum SGD using a constant learning rate.

Below are some critical lemmas for the proof of Theorem~\ref{theo:global}.
\begin{lemma}\label{lemma:seq}

Given two non-negative sequences $\{a_t\}_{t=1}^{\infty}$ and $\{b_t\}_{t=1}^{\infty}$ that satisfying
\begin{equation}
a_t =  \sum_{s=1}^t\rho^{t-s}b_{s}, \numberthis \label{eqn1}
\end{equation}
with $\rho\in[0,1)$, we have
\begin{align*}
D_k:=\sum_{t=1}^{k}a_t^2 \leq & % \frac{(1+\frac{2\rho}{1-\rho})\sum_{t=1}^kb_t^2-\rho^2a_k^2}{1-\rho^2} \\ \leq &
\frac{1}{(1-\rho)^2} \sum_{s=1}^kb_s^2.
\end{align*}
%where $S_k=\sum_{t=1}^{k}a_t$, and $D_k=\sum_{t=1}^{k}a_t^2$.
\end{lemma}

\begin{proof}
From the definition, we have
\begin{align*}
S_k= & \sum_{t=1}^{k}\sum_{s=1}^t\rho^{t-s}b_{s}
=  \sum_{s=1}^{k}\sum_{t=s}^k\rho^{t-s}b_{s}
=  \sum_{s=1}^{k}\sum_{t=0}^{k-s}\rho^{t}b_{s}
\leq  \sum_{s=1}^{k}{b_{s}\over 1-\rho}, \numberthis \label{eqn3}\\
D_k=  & \sum_{t=1}^{k}\sum_{s=1}^t\rho^{t-s}b_{s}\sum_{r=1}^t\rho^{t-r}b_{r}\\
= & \sum_{t=1}^{k}\sum_{s=1}^t\sum_{r=1}^t\rho^{2t-s-r}b_{s}b_{r} \\
\leq &  \sum_{t=1}^{k}\sum_{s=1}^t\sum_{r=1}^t\rho^{2t-s-r}{b_{s}^2+b_{r}^2\over2}\\
= & \sum_{t=1}^{k}\sum_{s=1}^t\sum_{r=1}^t\rho^{2t-s-r}b_{s}^2 \\
\leq  & {1\over 1-\rho}\sum_{t=1}^{k}\sum_{s=1}^t\rho^{t-s}b_{s}^2\\
\leq & {1\over (1-\rho)^2}\sum_{s=1}^{k}b_{s}^2, \quad \text{(due to \eqref{eqn3})}
\end{align*}
which completes the proof.
\end{proof}

\begin{lemma}\label{lemma:supp_main}
Under Assumption~\ref{ass:global}, for any sequence that follows the updating rule of
\begin{align*}
\x_{t+1} =& \x_t - \gamma \m_t\\
\m_t =& \beta\m_{t-1} + (1-\beta)\overline{\bm{g}}_t + \overline{\bm{\delta}}_{t-1} - \overline{\bm{\delta}}_{t},
\end{align*}
if 
\begin{align*}
\mathbb E \overline{\g}_t =  \nabla & f(\x_t),\quad\mathbb E \|\overline{\g}_t - \nabla f(\x_t)\|^2\leq \frac{\sigma^2}{n},\quad \mathbb E\|\overline{\bm{\delta}}_t\|^2\leq \epsilon^2,\quad \forall t,\\
&\|\nabla f(\bm{x}) - \nabla f(\bm{y}) \| \leq L \|\bm{x} - \bm{y} \|,\quad \forall \bm{x},\forall \bm{y},
\end{align*}
then we can guarantee that
\begin{align*}
&\left(1-\gamma L - \frac{2\gamma^2 L^2}{(1-\beta)^2} \right)\sum_{t=0}^T \mathbb E\|\nabla f(\bm{x}_t)\|^2\\
 \leq & \frac{2\mathbb E f(\bm{x}_{1}) - 2\mathbb Ef(\bm{x}^*)}{\gamma}   + \frac{6\gamma^2L^2\epsilon^2 T}{(1-\beta)^2}  +  \frac{L\gamma\sigma^2T}{n} + \frac{2\gamma^2L^2\sigma^2 T}{n(1-\beta)^2}
\end{align*}

\end{lemma}

\begin{proof}
Instead of investigating $\bm{x}_t$ directly, we introduce the following sequence
\begin{align*}
\bm{y}_t = \bm{x}_t - \frac{\gamma}{1-\beta}(\bm{m}_t + \overline{\bm{\delta}}_{t-1}) .
\end{align*}
The updating rule of $\bm{y}_t$ admits
\begin{align*}
\bm{y}_{t+1} - \bm{y}_t = & \bm{x}_{t+1} - \bm{x}_t - \frac{\gamma}{1- \beta}(\bm{m}_{t+1} - \bm{m}_t - \overline{\bm{\delta}}_{t+1} + \overline{\bm{\delta}}_t) \\
= & -\gamma \bm{m}_t  - \frac{\gamma}{1-\beta}(\beta \bm{m_t} + (1-\beta)\bm{g}_t + \overline{\bm{\delta}}_{t-1} - \overline{\bm{\delta}}_{t} - \bm{m}_t + \overline{\bm{\delta}}_{t} - \overline{\bm{\delta}}_{t-1})\\
= & -\gamma\bm{g}_t.
\end{align*}
Since $f(\cdot)$ is with L-Lipschitzian, we have
\begin{align*}
\mathbb E f(\bm{y}_{t+1}) - \mathbb Ef(\bm{y}_{t}) \leq & \mathbb E\left\langle \nabla f(\bm{y}_t), \bm{y}_{t+1} - \bm{y}_t \right\rangle + \frac{L}{2}\mathbb E\left\|\bm{y}_{t+1} - \bm{y}_t \right\|^2\\
=& -\gamma \mathbb E\left\langle \nabla f(\bm{y}_t), \bm{g}_t\right\rangle + \frac{L\gamma^2}{2}\mathbb E\|\bm{g}_t\|^2\\
=& -\gamma \mathbb E\left\langle \nabla f(\bm{y}_t), \nabla f(\bm{x}_t)\right\rangle + \frac{L\gamma^2}{2}\mathbb E\|\bm{g}_t\|^2\\
= & -\frac{\gamma}{2} \mathbb E\|\nabla f(\bm{x}_t)\|^2 - \frac{\gamma}{2}\mathbb E\|\nabla f(\bm{y}_t)\|^2 + \frac{\gamma}{2} \mathbb E\|\nabla f(\bm{x}_t) - \nabla f(\bm{y}_t)\|^2+ \frac{L\gamma^2}{2}\mathbb E\|\bm{g}_t\|^2\\
\leq & -\frac{\gamma}{2} \mathbb E\|\nabla f(\bm{x}_t)\|^2 + \frac{\gamma L^2}{2}\mathbb E\|\bm{x}_t - \bm{y}_t \|^2+ \frac{L\gamma^2}{2}\mathbb E\|\bm{g}_t\|^2\\
= & -\frac{\gamma}{2} \mathbb E\|\nabla f(\bm{x}_t)\|^2 + \frac{\gamma^3 L^2}{2}\mathbb E\left\|\frac{\bm{m}_t}{1-\beta} + \frac{\overline{\bm{\delta}}_{t-1}}{1-\beta} \right\|^2+ \frac{L\gamma^2}{2}\mathbb E\|\bm{g}_t\|^2\\
\leq & -\frac{\gamma}{2} \mathbb E\|\nabla f(\bm{x}_t)\|^2 + \frac{\gamma^3 L^2}{(1-\beta)^2}\mathbb E\|\bm{m}_t\|^2 + \frac{\gamma^3L^2}{(1-\beta)^2}\mathbb E\|\overline{\bm{\delta}}_{t-1}\|^2  +  \frac{L\gamma^2}{2}\mathbb E\|\bm{g}_t\|^2\\
\leq & -\frac{\gamma}{2} \mathbb E\|\nabla f(\bm{x}_t)\|^2 + \frac{\gamma^3 L^2}{(1-\beta)^2}\mathbb E\|\bm{m}_t\|^2 + \frac{\gamma^3L^2\epsilon^2}{(1-\beta)^2}  +  \frac{L\gamma^2}{2}\mathbb E\|\bm{g}_t\|^2\\
\leq & -\frac{\gamma}{2} \mathbb E\|\nabla f(\bm{x}_t)\|^2 + \frac{\gamma^3 L^2}{(1-\beta)^2}\mathbb E\|\bm{m}_t\|^2 + \frac{\gamma^3L^2\epsilon^2}{(1-\beta)^2}  +  \frac{L\gamma^2}{2}\mathbb E\|\nabla f(\bm{x}_t)\|^2 + \frac{L\gamma^2\sigma^2}{2n}.
\end{align*}
Summing up the equation above from $t=0$ to $t=T$ we get
\begin{align*}
\mathbb E f(\bm{y}_{T+1}) - \mathbb Ef(\bm{y}_{0}) \leq -\frac{(1-\gamma L)\gamma}{2}\sum_{t=0}^T \mathbb E\|\nabla f(\bm{x}_t)\|^2 + \frac{\gamma^3 L^2}{(1-\beta)^2}\sum_{t=0}^T\mathbb E\|\bm{m}_t\|^2 + \frac{\gamma^3L^2\epsilon^2 T}{(1-\beta)^2}  +  \frac{L\gamma^2\sigma^2 T}{2n},
\end{align*}
which can be rewritten into
\begin{align*}
(1-\gamma L)\sum_{t=0}^T \mathbb E\|\nabla f(\bm{x}_t)\|^2 \leq & \frac{2\mathbb E f(\bm{y}_{0}) - 2\mathbb Ef(\bm{y}_{T+1})}{\gamma}  +  \frac{2\gamma^2 L^2}{(1-\beta)^2}\sum_{t=0}^T\mathbb E\|\bm{m}_t\|^2 + \frac{2\gamma^2L^2\epsilon^2 T}{(1-\beta)^2}  +  \frac{L\gamma\sigma^2 T}{n}.\numberthis\label{supp:final_eq1}
\end{align*}

Notice that we have
\begin{align*}
\bm{m}_t =& (1-\beta) \sum_{s=0}^t \beta^{t-s} \overline{\bm{g}}_s + \sum_{s=0}^t \beta^{t-s}( \overline{\bm{\delta}}_{s-1} - \overline{\bm{\delta}}_{s})
\end{align*}
which by using Lemma~\ref{lemma:seq}, we have
\begin{align*}
\sum_{t=0}^T\|\bm{m}_t\|^2 \leq \sum_{t=0}^T \|\bm{g}_t\|^2 + \frac{2}{(1-\beta)^2}\sum_{t=0}^T \|\overline{\bm{\delta}}_t\|^2 \leq \sum_{t=0}^T \|\nabla f(\bm{x}_t)\|^2 + \frac{\sigma^2T}{n} + \frac{2\epsilon^2T}{(1-\beta)^2} . \numberthis\label{supp:final_nound_mt}
\end{align*}
Combing \eqref{supp:final_eq1} and \eqref{supp:final_nound_mt} together we get
\begin{align*}
&\left(1-\gamma L - \frac{2\gamma^2 L^2}{(1-\beta)^2} \right)\sum_{t=0}^T \mathbb E\|\nabla f(\bm{x}_t)\|^2\\
 \leq & \frac{2\mathbb E f(\bm{y}_{0}) - 2\mathbb Ef(\bm{y}_{T+1})}{\gamma}   + \frac{6\gamma^2L^2\epsilon^2 T}{(1-\beta)^2}  +  \frac{L\gamma\sigma^2T}{n} + \frac{2\gamma^2L^2\sigma^2 T}{n(1-\beta)^2}\\
 \leq & \frac{2\mathbb E f(\bm{x}_{1}) - 2\mathbb Ef(\bm{x}^*)}{\gamma}   + \frac{6\gamma^2L^2\epsilon^2 T}{(1-\beta)^2}  +  \frac{L\gamma\sigma^2T}{n} + \frac{2\gamma^2L^2\sigma^2 T}{n(1-\beta)^2}.
\end{align*}
\end{proof}

\paragraph{Proof to Theorem~\ref{theo:global}} Since using a per-coordinate learning rate for loss function $f(\cdot)$ is equivalent to use a constant learning for all coordinates but for loss function $h(\cdot)$, the only two thing that change are
\begin{itemize}
\item \textbf{Different L-Lipschitzian coefficient}: the L-Lipschitzian coefficient for $h(\cdot)$ is
\begin{align*}
\|\nabla h(\x) - \nabla h(\bm{y})\|^2 =& \left\|V^{\frac{1}{2}} \nabla f(V^{\frac{1}{2}}\x) - V^{\frac{1}{2}} \nabla f(V^{\frac{1}{2}}\bm{y}) \right\|^2 \\
= & \left\| \nabla f(V^{\frac{1}{2}}\x) - \nabla f(V^{\frac{1}{2}}\bm{y}) \right\|^2_V\\
\leq & L^2\left\| V^{\frac{1}{2}}\x - V^{\frac{1}{2}}\bm{y} \right\|^2_V\\
=& L^2\|\x - \bm{y}\|^2_{V^2}\\
\leq & L^2V_{\max}^2\|\x - \bm{y}\|^2.
\end{align*}
Therefore the effective L-Lipschitzian coefficient of $h(\x)$ is $LV_{\max}$
\item \textbf{Different definition of $\overline{\bm{\delta}}_t$}: from \eqref{supp:trans_eq1} we shall see that actually the compression error in the view of $h(\cdot)$ is $V^{\frac{1}{2}}\overline{\bm{\delta}}_t $, so in this case we have
\begin{align*}
\mathbb E\|V^{\frac{1}{2}}\overline{\bm{\delta}}_t\|^2 \leq V_{\max}\epsilon^2
\end{align*}

\end{itemize}
\begin{proof}
From Lemma~\ref{lemma:supp_main}, we have
\begin{align*}
&\left(1-\gamma L - \frac{2\gamma^2 L^2V_{\max}^2}{(1-\beta)^2} \right)\sum_{t=0}^T \mathbb E\|\nabla h(\bm{z}_t)\|^2\\
 \leq & \frac{2\mathbb E f(\bm{x}_{0}) - 2\mathbb Ef(\bm{x}^*)}{\gamma}   + \frac{6\gamma^2L^2\epsilon^2V_{\max}^3 T}{(1-\beta)^2}  +  \frac{L\gamma V_{\max}\sigma^2T}{n} + \frac{2\gamma^2L^2\sigma^2 V_{\max}^2 T}{n(1-\beta)^2},
\end{align*}
which by using $\nabla h(\bm{z}_t) = V^{\frac{1}{2}}\nabla f(\x_t) $, it becomes
\begin{align*}
&\left(1-\gamma LV_{\max} - \frac{2\gamma^2 L^2V_{\max}^2}{(1-\beta)^2} \right)\sum_{t=0}^T \mathbb E\|\nabla f(\bm{x}_t)\|^2_{V}\\
 \leq & \frac{2\mathbb E f(\bm{x}_{0}) - 2\mathbb Ef(\bm{x}^*)}{\gamma}   + \frac{6\gamma^2L^2\epsilon^2V_{\max}^3 T}{(1-\beta)^2}  +  \frac{L\gamma V_{\max}\sigma^2T}{n} + \frac{2\gamma^2L^2\sigma^2 V_{\max}^2 T}{n(1-\beta)^2},
\end{align*}
Since $V_{\max} = \frac{1}{\sqrt{v_{\min}}}$, therefore the equation above becomes
\begin{align*}
  &\left(1-\frac{\gamma L}{v_{\min}} - \frac{2\gamma^2 L^2}{(1-\beta)^2v_{\min}^2} \right)\sum_{t=0}^T \mathbb E\|\nabla f(\bm{x}_t)\|^2_{V}\\
  \leq & \frac{2\mathbb E f(\bm{x}_{0}) - 2\mathbb Ef(\bm{x}^*)}{\gamma}   + \frac{6\gamma^2L^2\epsilon^2 T}{(1-\beta)^2v_{\min}^3}  +  \frac{L\gamma \sigma^2T}{nv_{\min}} + \frac{2\gamma^2L^2\sigma^2  T}{n(1-\beta)^2v_{\min}^2},
\end{align*}
\end{proof}

\section{Proof to Corollary~\ref{coro:global}}
\begin{proof}
By choosing $\gamma =\frac{1-\beta}{4LV_{\max} + \sigma\sqrt{\frac{T}{n}} + T^{^{\frac{1}{3}}}\epsilon^{^{\frac{2}{3}}} }   $, we can guarantee that
\begin{align*}
1- \gamma L - \frac{2\gamma^2 L^2V_{\max}^2}{(1-\beta)^2} \geq & \frac{1}{2}.
\end{align*}
So \eqref{main:theo:eq} leads to
\begin{align*}
\sum_{t=0}^T \mathbb E\|\nabla f(\bm{x}_t)\|^2_V \leq & \frac{2\left(\mathbb E f(\bm{y}_{0}) - f(\bm{y}^*) \right)}{(1-\beta)}\left(4LV_{\max} + \sigma\sqrt{\frac{T}{n}} + T^{^{\frac{1}{3}}}\epsilon^{^{\frac{2}{3}}} \right)\\
& + \left((1-\beta)L\sqrt{T}+ 2L^2V_{\max}^2\right)\frac{\sigma}{\sqrt{n}} +   6L^2\epsilon^{\frac{2}{3}}T^{\frac{1}{3}}V_{\max}^3\\
\frac{1}{T}\sum_{t=0}^T \mathbb E\|\nabla f(\bm{x}_t)\|^2_V \leq & \frac{2\left(\mathbb E f(\bm{y}_{0}) - f(\bm{y}^*) \right)}{(1-\beta)}\left(\frac{4LV_{\max}}{T} + \frac{\sigma}{\sqrt{nT}} + T^{^{-\frac{2}{3}}}\epsilon^{^{\frac{2}{3}}} \right)\\
& + \left((1-\beta)L+ \frac{2L^2V_{\max}^2}{\sqrt{T}}\right)\frac{\sigma}{\sqrt{nT}} +   6L^2\epsilon^{\frac{2}{3}}T^{-\frac{2}{3}}V_{\max}^3.
\end{align*}
Treating $f(\bm{y}_1) - f^*$, $\beta$ and $L$ as constants, from the inequality above we get
\begin{align*}
\frac{1}{T}\sum_{t=0}^T \mathbb E\|\nabla f(\bm{x}_t)\|^2 \lesssim \frac{\sigma}{\sqrt{nT}} + \frac{\epsilon^{\frac{2}{3}}}{T^{\frac{2}{3}}} + \frac{1}{T}.
\end{align*}
It completes the proof.
\end{proof}

\end{document}